\documentclass[10pt,twocolumn,letterpaper]{article}

\usepackage{wacv}
\usepackage{times}
\usepackage{epsfig}
\usepackage{graphicx}
\usepackage{amssymb}
\usepackage{amsmath}
\usepackage{amsmath }
\usepackage{authblk}
\usepackage{multirow}
\usepackage{hhline}
\usepackage{algorithm}
\usepackage{algpseudocode}
\usepackage[font=footnotesize,skip=0pt]{caption}
\usepackage{array}
\usepackage{empheq}
\usepackage{graphicx}
\graphicspath{{images_jpg/}}
\usepackage{amsmath}
\DeclareMathOperator*{\argmax}{arg\,max}

\wacvfinalcopy 

\def\wacvPaperID{288} 

\begin{document}

\title{Spacetime Graph Optimization for Video Object Segmentation}

\author[1,2,3]{Emanuela Haller}
\author[1]{Adina Magda Florea}
\author[1,2,3]{Marius Leordeanu}
\affil[1]{University Politehnica of Bucharest}
\affil[2]{Institute of Mathematics of the Romanian Academy}
\affil[3]{Bitdefender, Romania}
\affil[ ]{\small{\tt {haller.emanuela@gmail.com  adinamagdaflorea@yahoo.com  leordeanu@gmail.com}}}

\maketitle

\begin{abstract}

 We address the challenging task of foreground object discovery and segmentation in video. We introduce an efficient solution, suitable for both unsupervised and supervised scenarios, based on a spacetime graph representation of the video sequence. We ensure a fine grained representation with one-to-one correspondences between graph nodes and video pixels. We formulate the task as a spectral clustering problem by exploiting the spatio-temporal consistency between the scene elements in terms of motion and appearance. Graph nodes that belong to the main object of interest should form a strong cluster, as they are linked through long range optical flow chains and have similar motion and appearance features along those chains. On one hand, the optimization problem aims to maximize the segmentation clustering score based on the motion structure through space and time. On the other hand, the segmentation should be consistent with respect to node features. Our approach leads to a graph formulation in which the segmentation solution becomes the principal eigenvector of a novel Feature-Motion matrix. While the actual matrix is not computed explicitly, the proposed algorithm efficiently computes, in a few iteration steps, the principal eigenvector that captures the segmentation of the main object in the video. The proposed algorithm, GO-VOS, produces a global optimum solution and, consequently, it does not depend on initialization. In practice, GO-VOS achieves state of the art results on three challenging datasets used in current literature: DAVIS, SegTrack and YouTube-Objects.
\end{abstract}

\section{Introduction}
Discovering objects in videos without human supervision, as they move and change appearance over space and time, is one of the most challenging, and still unsolved problems in computer vision. This impacts the way we learn about objects and how we process large amounts of video data, which is widely available at low cost.  One of our core goals is to understand how much we could learn automatically from a video about the main object of interest. We are interested in how object properties relate in both space and time and how we could exploit these consistencies in order to discover the objects in a fast and accurate manner. While human segmentation annotations of the same video are not always in agreement, people tend to agree on which is the main object captured in a given video shot, when the camera is indeed focusing on a single object. There are many questions to answer: which are the particular features that make a group of pixels stand out as a single, main object in a given sequence? Is it the pattern of motion different from the surrounding background? Is it the contrast in appearance, the symmetry or good form of an object that make it stand out? Or is it a combination of such factors and maybe others, still unknown? What we do know is the fact that people can easily detect and segment the foreground object when the camera is focusing on a single object, and this fact has been recognized and studied since the early days of the Gestalt school of psychology \cite{koffka2013principles}.

We make two main assumptions, which become the basis of our approach: \textbf{1)}~pixels that belong to the same object are highly likely to be connected through long range optical flow chains, as we define them in Sec. \ref{sec:approach_space_time_graph}; \textbf{2)}~pixels of the same object are also likely to have similar motion and distinctive appearance patterns in space and time. In other words, what looks alike and moves together, is likely to belong together. While these ideas are not new, we propose a novel graph structure in space and time, with motion and appearance constraints, in which segmentation is defined as a clustering problem 
in the spacetime graph, in which the global optimum is found as the 
the leading eigenvector of a novel Feature-Motion matrix (Sec. \ref{sec:alg_analysis_convergence}).


\textbf{Main contribution:} We introduce a novel graph structure in space and time, with nodes at the dense pixel level such that each pixel is connected to other pixels through optical flow chains. The graph combines long range motion patterns with local appearance information at the dense pixel level, in a single spacetime graph defined by the Feature-Motion matrix (Sec \ref{sec:alg_analysis_convergence}). We define segmentation (Problems 1 and 2 - Eq.~\ref{optimization_problem} and Eq.~\ref{optimization_problem_alg}) as a spectral clustering problem and
propose a fast algorithm that computes the global optimum as the principal eigenvector of the Feature-Motion matrix. One of the main tricks is that the matrix is never computed explicitly, such that the algorithm is fast and also accurate, with state of the art performance on three challenging benchmarks in the literature.

\textbf{Related work:} One of the most important aspects that differentiates between different approaches in foreground object segmentation in video is the amount of supervision used, which could vary from complete absence of any human annotations \cite{lao2018extending, papazoglou2013fast, keuper2015motion, faktor2014video, haller2017unsupervised} to using models pretrained for video object segmentation \cite{luiten2018premvos, maninis2017video, voigtlaender2017online, bao2018cnn, wug2018fast, cheng2018fast, caelles2017one, perazzi2017learning, chen2018blazingly, song2018pyramid, tokmakov2017learning, jain2017fusionseg} or having access to the human annotation for the first video frame \cite{Perazzi2016}. 
We propose a video object segmentation framework that has the ability to accommodate both supervised and unsupervised scenarios. In this paper we focus on the unsupervised case, for which no human supervision was used during training. 

Our approach uses optical flow in order to define the graph structure at the dense pixel level. There are several notable works that use optical flow or a graph representation for tasks related to video object segmentation (VOS). Brox and Malik introduce in \cite{brox2010object} a motion clustering method that simultaneous estimates optical flow and object segmentation masks. They operate on pairs of images and minimize an energy function incorporating classical optical flow constrains and a descriptor matching term, with no mechanism for selecting the main object. Another similar approach is proposed by Tsai \etal in \cite{tsai2016video}. Zhuo \etal \cite{zhuo2018unsupervised} build salient motion masks that are further combined with "objectness" masks in order to generate the final segmentation. Li \etal \cite{li2017primary} and Wang \etal \cite{wang2015saliency} introduce approaches that are based on averaging saliency masks over connections defined by optical flow. Keuper \etal introduce in \cite{keuper2015motion} a method for motion segmentation, where elements of the scene are clustered function of their motion patterns, with no focus on the main object. They formulate the task as a minimum cost multicut problem over a graph whose nodes are defined by motion trajectories. 

Our spacetime graph considers direct connections between all video pixels, in contrast to superpixel level nodes or trajectory nodes. In contrast to \cite{keuper2015motion, brox2010object} that perform motion segmentation, we discover the strongest cluster in space and time by taking in consideration both long range motion connections as well as local motion and appearance patterns. In contrast to other works in VOS, we provide a spectral clustering formulation with global solution that can be obtained very fast by power iteration as the leading eigenvector of a huge Feature-Motion matrix (of size $n \times n$, where $n$ is the total number of pixels in the video), which is never explicitly computed. Thus we ensure a dense consistency in space and time at the level of the whole video sequence, in contrast to local consistency that is ensured in most existing solutions.


\begin{figure*}[!h]    
        \includegraphics[width=\textwidth]{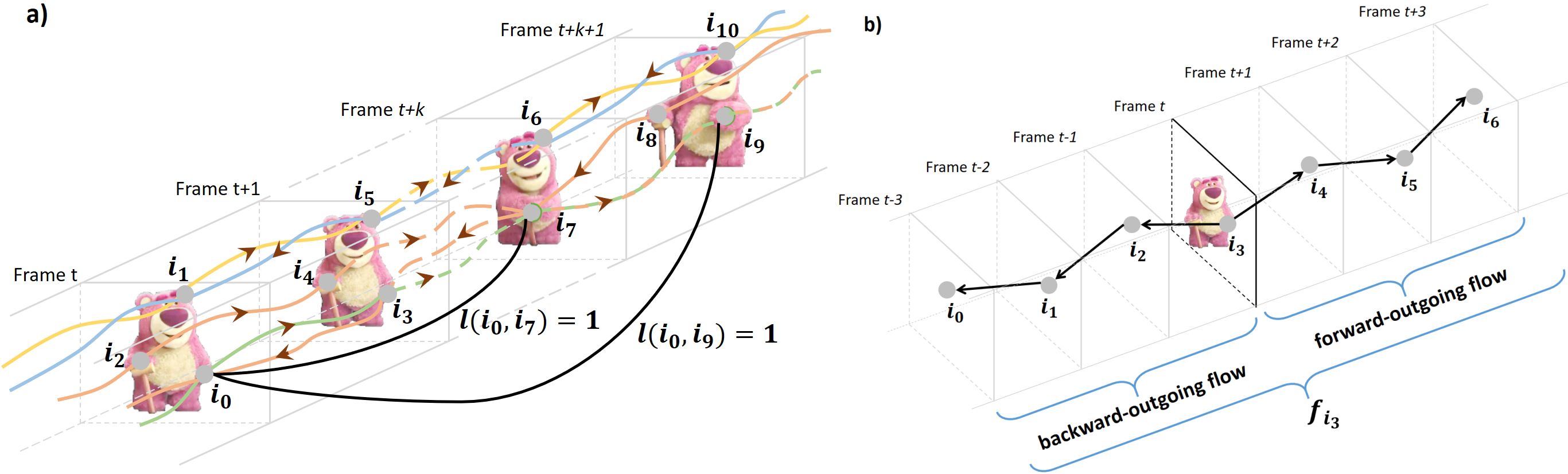}
        \centering
        \caption{Visual representation of how the spacetime graph is formed. \textbf{a)} $i_0$ and $i_7$ are pixels in different frames, which become nodes in the graph. A flow chain connects points that are linked by optical flow in the same direction (forward or backward) along a path of flows between consecutive frames (colored lines). Through a given node there are always two chains going out, one in each direction. However, there might be several chains or none coming in from both directions (e.g. node $i_7$). So, (out)degree of a node $=2$, (in)degree $\geq 0$. Thus, there could be multiple flow chains connecting any two nodes (pixels) in the graph. The flow chains define a graph structure, in which pixels are connected if there is at least one flow chain between them (black lines correspond to graph edges). One of the assumptions we make is that the stronger a cluster, the more likely it belongs to the main  object. Thus, pixels that are strongly connected to other foreground pixels, are also likely to be foreground themselves. \textbf{b)} Flow chains are also used to describe node features, by providing patterns of motion (vectors of optical flow displacements) and patterns of appearance and other pre-computed features along the chains. For a given node $i_3$, the features forming $\mathbf{f}_{i_3}$ are collected along the outgoing flow chains.}
        \label{fig:space_time_graph}
        \vspace{-2.5mm}
    \end{figure*}
    
\section{Our approach}\label{sec:approach}
    Given a sequence of $m$ consecutive video frames, our aim is to extract a set of $m$ soft-segmentation masks, one per frame, containing the main object of interest. We represent the entire video as a graph with one node per pixel and a structure defined by optical flow chains, as shown in Sec.~\ref{sec:approach_space_time_graph}. We formulate segmentation as a clustering problem in Secs.~\ref{sec:approach_problem_formulation} and \ref{sec:alg_analysis_convergence}. We provide a first formulation for segmentation in Problem 1 (Eq.~\ref{optimization_problem}), for which we derive an analytical solution in Sec.~\ref{sec:approach_optimization_problem}. Since the initial Problem 1 is intractable in practice, we provide an efficient algorithm in Sec.~\ref{sec:approach_algorithm_implementation}, which optimally solves a more tractable task (Problem 2 (Eq.~\ref{optimization_problem_alg}) defined in Sec. \ref{sec:alg_analysis_convergence}), which is a slightly modified version of Problem 1.
    
\subsection{Spacetime graph}\label{sec:approach_space_time_graph}
    
    \textbf{Graph of pixels in spacetime:} in the spacetime graph $G=(V,E)$, each node $i\in V$ is associated to a pixel in one of the video frames. $V$ has $n$ nodes, with $n=m\cdot h\cdot w$, where $(h,w)$ is the frame size and $m$ the number of frames.
    
    \textbf{Optical flow chains:} given optical flows between pairs of consecutive frames, both forward and backward, we form optical flow chains by following the flow (in the same direction) between consecutive frames, starting from a given pixel in a frame, then go frame by frame, all the way to the end of the video. Thus, through a pixel could pass multiple chains, at least one moving forward and one moving backward. A chain in a given direction could start at that pixel, if there is no incoming flow in that direction or it could pass through that pixel and start at a different previous frame w.r.t to that particular direction. Note that, for a given direction, a pixel could have none or several incoming flows, whereas it will always have only one outgoing flow chain. These flow chains are important in our graph as they define its edges. Thus, there is an undirected edge between two nodes $(i,j)$ if they are connected by an optical flow chain in one direction or both. Note that based on our definition above, there could be a maximum of two different optical flow chains between two nodes, one per direction. 
    
    \textbf{Adjacency matrix:} we introduce the adjacency matrix $\mathbf{M}\in \mathbb{R}^{n \times n}$, defined as $\mathbf{M}_{i,j} = l(i,j) \cdot k(i,j)$, where $k(i,j)$ is a Gaussian kernel as a function of the temporal distance between nodes $i$ and $j$, while $l(i,j)=1$ if there is an edge between $i$ and $j$ and zero otherwise. Thus, $\mathbf{M}_{i,j} = k(i,j)$ if $i$ and $j$ are connected and zero otherwise. According to the definition, $\mathbf{M}$ is also symmetric, semi-positive definite, has non-negative elements and is expected to be very sparse. $\mathbf{M}$ is a Mercer kernel, since the pairwise terms $\mathbf{M}_{i,j}$ satisfy Mercer's condition. In Figure~\ref{fig:space_time_graph}.a, we introduce a visual representation of how edges in the spacetime graph are formed through long range optical flow chains.
    
    \textbf{Nodes labels and their features:} besides the graph structure, completely described by pairwise functions between nodes, in $\mathbf{M}$, each node $i$ is also described by unary, node-level feature vectors $\mathbf{f}_i$, collected along the two outgoing chains starting at $i$, one per direction (Figure~\ref{fig:space_time_graph}.b). We stack all features into a feature matrix $\mathbf{F} \in  \mathbb{R}^{n \times d}$.
    In practice we can consider different features, pretrained or not, but we refer the reader to Sec.~\ref{sec:approach_algorithm_implementation} and Sec.~\ref{sec:alg_analysis_features} for details. 
    
    Each node has a (soft) segmentation label $x_i \in [0,1]$, which, at any moment in time represents our belief that the node is part of the object of interest. Thus, we can represent a solution to the segmentation problem, over the whole video, as a vector of labels $\mathbf{x} \in \mathbb{R}^{n \times 1}$, with a label $x_i$ for each pixel $i$. The second assumption we make is that nodes with similar features should have similar labels. From a mathematical point of view, we want to be able to regress the labels $x_i$ on the features $\mathbf{f}_i$ - this says that the features associated with a node should suffice for predicting its label. If the regression is possible with sufficiently small error, then the assumption that pixels with similar features have similar labels is automatically satisfied. 

    Now, we are prepared to formulate the segmentation problem mathematically. On one hand we want to find a strong cluster in $\mathbf{M}$, as defined by $\mathbf{x}$, on the other we want to be able to regress $\mathbf{x}$ on the node features $\mathbf{F}$. In the next section we show that these factors interact and define object segmentation in video as an optimization problem.
    
\subsection{Problem formulation}\label{sec:approach_problem_formulation}
    
    Nodes belonging to the main object of interest should form a strong cluster in the spacetime graph, such that they are strongly connected through long range flow chains and their features are able to predict their labels $\mathbf{x}$.  Vector $\mathbf{x}$ represents the segmentation labels of individual nodes and also defines the segmentation cluster. Nodes with label 1 are part of this cluster, those with label zero are not. We define the intra-cluster score to be $S_C=\sum_{i,j\in V}\mathbf{x}_i \mathbf{x}_j \mathbf{M}_{i,j}$, which can be written in the matrix form as  $S_C(\mathbf{x})=\mathbf{x}^{T}  \mathbf{M}  \mathbf{x}$.
    
    We relax the condition on $\mathbf{x}$, allowing continuous values for the labels in $\mathbf{x}$. For the purpose of video object segmentation, we only care about the labels' relative values, so for stability of convergence, we impose the L2-norm of vector $\mathbf{x}$ to be 1. The pairwise links $\mathbf{M}_{i,j}$ are stronger when nodes $i$ and $j$ are linked through flow chains and close to each other, therefore we want to maximize the clustering score. Under the constraint $\|\mathbf{x}\|_2=1$, the score $\mathbf{x}^{T}  \mathbf{M}  \mathbf{x}$ is maximized by the leading eigenvector of $\mathbf{M}$, which must have non-negative values by Perron-Frobenius theorem, since the matrix has non-negative elements. Finding the main cluster by inspecting the main eigenvector of the adjacency matrix is a classic case of spectral clustering~\cite{meila_shi} and also related to spectral approaches in graph matching~\cite{leordeanu2012unsupervised}. However, in our case $\mathbf{M}$ alone is not sufficient for our problem, since it is defined by simple connections between nodes with no information regarding their appearance or higher level features to better capture their similarity.
    
    As mentioned, we impose the constraint that nodes having similar features should have similar labels. We require that $\mathbf{x}$ should be predicted from the features $\mathbf{F}$, through a linear mapping: $\mathbf{x} \sim \mathbf{F} \mathbf{w}$, for some $\mathbf{w} \in \mathbb{R}^{d \times 1}$.  Thus, besides the problem of maximizing the clustering score $\mathbf{x}^{T}  \mathbf{M}  \mathbf{x}$, we also aim to minimize an error term  ${\|\mathbf{F}\mathbf{w}-\mathbf{x}\|}_2$, which enforces a feature-label consistency such that labels could be predicted well from features. After including a regularization term ${\|\mathbf{w}\|}_2$ which should be minimized, we obtain the final objective score for segmentation:
    
    \begin{equation}
        \label{objective_function}
        S(\mathbf{x}, \mathbf{w})=\mathbf{x}^{T}\mathbf{M} \mathbf{x} - \alpha (\mathbf{F}\mathbf{w}-\mathbf{x})^{T}(\mathbf{F}\mathbf{w}-\mathbf{x})-\beta\mathbf{w}^{T}\mathbf{w}.
    \end{equation}
    
    \noindent To find a solution we should maximize this objective subject to the constraint $\|\mathbf{x}\|_2=1$, resulting in the optimization problem:
    
    \begin{equation}
        \label{optimization_problem}
        \textbf{Problem 1:} \quad (\mathbf{x}^{*}, \mathbf{w}^{*}) = \argmax_{\mathbf{x},\mathbf{w}} S(\mathbf{x},\mathbf{w}) \quad \text{s.t.} \quad \|\mathbf{x}\|_2=1
    \end{equation}
    
\subsection{Finding the optimal segmentation}\label{sec:approach_optimization_problem}
    The optimization problem defined in Eq.~\ref{optimization_problem} requires that we find a maxima of the function $S(\mathbf{x}, \mathbf{w})$, subject to an equality constraint $\mathbf{x}^{T}\mathbf{x} = 1$. In order to solve this problem, we introduce the Lagrange multiplier $\lambda$ and define the Lagrange function:
    
    \begin{equation}
        \label{lagrange_function}
         \mathcal{L}(\mathbf{x}, \mathbf{w}, \mathbf{\lambda})=S(\mathbf{x},\mathbf{w})-\lambda(\mathbf{x}^{T}\mathbf{x}-1).
    \end{equation}

    The stationary points satisfy $\nabla_{\mathbf{x}, \mathbf{w}, \lambda} \mathcal{L}(\mathbf{x}, \mathbf{w}, \lambda)=\mathbf{0}$. By the end of this section we will show that the stationary point is also a global optimum of our problem (Eq.~\ref{optimization_problem}), the principal eigenvector of a specific matrix $\mathbf{A}$ (Eq.~\ref{power_iteration_th}). Next, we obtain the following system of equations:
    
    \begin{equation}
        \label{lagrange_equations}
        \begin{cases}
            \nabla_\mathbf{x}\mathcal{L}(\mathbf{x}, \mathbf{w}, \lambda)=\mathbf{0} \Rightarrow \mathbf{x}(-\lambda-\alpha)+\mathbf{M}\mathbf{x}+2\alpha\mathbf{F}\mathbf{w} = \mathbf{0} \\
            \nabla_\mathbf{w}\mathcal{L}(\mathbf{x}, \mathbf{w}, \lambda)=\mathbf{0} \Rightarrow \mathbf{w}(\alpha\mathbf{F}^{T}\mathbf{F}+\beta\mathbf{I}_d)-2\alpha\mathbf{F}^{T}\mathbf{x}=\mathbf{0} \\ 
            \nabla_\lambda\mathcal{L}(\mathbf{x}, \mathbf{w}, \lambda)=\mathbf{0}  \Rightarrow \mathbf{x}^{T}\mathbf{x}-1=0
        \end{cases}
    \end{equation}
    
    From $\nabla_\mathbf{w}\mathcal{L}(\mathbf{x}, \mathbf{w}, \lambda)=\mathbf{0}$ we arrive at the closed-form solution for ridge regression, with optimum $\mathbf{w}^*(\mathbf{x})=2\alpha{(\alpha\mathbf{F}^{T}\mathbf{F}+\beta\mathbf{I}_d)}^{-1}\mathbf{F}^{T}\mathbf{x}$, as a function of $\mathbf{x}$. All we have to do now is compute the optimum $\mathbf{x}^*$, for which we take a fixed-point iteration approach, which (as discussed in more detail later) should converge to a  solution of the equation $\nabla_\mathbf{x}\mathcal{L}(\mathbf{x}, \mathbf{w(x)}, \lambda)=0$. 
    
    We rewrite $\nabla_\mathbf{x}\mathcal{L}(\mathbf{x}, \mathbf{w(x)}, \lambda)=\mathbf{0}$ and  $\mathbf{x}^{T}\mathbf{x}-1=0$ in the form $\mathbf{x}=h(\mathbf{x})$ such that any fixed point of $h$ will be a solution for our initial equation. Thus, we apply a fixed point iteration scheme and iteratively update the value of $\mathbf{x}$ as a function of its previous value. We immediately obtain $h(\mathbf{x})=\frac{1}{p^{'}(-\lambda-\alpha)}\mathbf{M}\mathbf{x}+\frac{2\alpha}{p^{'}(-\lambda-\alpha)}\mathbf{F}\mathbf{w}^*(\mathbf{x})$, where $p^{'}=\|\frac{1}{-\lambda-\alpha}\mathbf{M}\mathbf{x}+\frac{2\alpha}{-\lambda-\alpha}\mathbf{F}\mathbf{w}\|_2$. The term $(-\lambda-\alpha)$ cancels out and we end up with the following power iteration scheme that optimizes the segmentation objective $S(\mathbf{x},\mathbf{w})$ (Eq. \ref{objective_function}) under L2 constraint $\mathbf{x}^T\mathbf{x}=1$: 

    \begin{equation}
        \label{theoretical_iterations}        
        \begin{cases}
        \mathbf{x}^{(it+1)}=\frac{1}{p}\mathbf{M}\mathbf{x}^{(it)}+\frac{2\alpha}{p}\mathbf{F}\mathbf{w}^{(it)} \\
        \mathbf{w}^{(it+1)}=2\alpha{(\alpha\mathbf{F}^{T}\mathbf{F}+\beta\mathbf{I}_d)}^{-1}\mathbf{F}^{T}\mathbf{x}^{(it+1)} \\
        \hfill\\
        \hfill p=\|\mathbf{M}\mathbf{x}^{(it)}+2\alpha\mathbf{F}\mathbf{w}^{(it)}\|_2,
        \end{cases}
    \end{equation}
  
    \noindent where $\mathbf{x}^{(it)}$ and $\mathbf{w}^{(it)}$ are the values of $\mathbf{x}$, respective $\mathbf{w}$ at iteration $it$. We have reached a set of compact segmentation updates at each iteration step, which efficiently combines the clustering score and the regression loss, while imposing a constraint on the norm of $\mathbf{x}$. Note that the actual norm is not important. Values in $\mathbf{x}$ are always non-negative and only their relative values matter. They can be easily scaled and shifted to range between 0 and 1 (without changing the direction of vector $\mathbf{x}$),
    which we actually do in practice.
    
    We can show that the stationary point of our optimization problem is in fact a global optimum, namely the principal eigenvector of a specific symmetric matrix, which we construct below. In Eq.~\ref{theoretical_iterations}, if we write $\mathbf{w}^{(it+1)}$ in terms of $\mathbf{x}^{(it+1)}$ and replace it in all equations, we can then write $\mathbf{x}^{(it+1)}$ as a function of $\mathbf{x}^{(it)}$, $\mathbf{M}$ and $\mathbf{F}$:
    
    \begin{equation}
        \label{power_iteration_th}
        \mathbf{x}^{(it+1)} = \frac{\mathbf{A}\mathbf{x}^{(it)}}{\|\mathbf{A}\mathbf{x}^{(it)}\|_2}, \\
    \end{equation}
    
    where $\mathbf{A}=\mathbf{M}+4\alpha^2\mathbf{F}(\alpha\mathbf{F}^{T}\mathbf{F}+\beta\mathbf{I}_d)^{-1}\mathbf{F}^{T}$. This solution is hard to obtain in practice due to the difficulty of working explicitly with matrix A (which is impossible to build). Next we present an efficient optimization algorithm that overcomes this limitation and solves optimally a slightly modified version of Problem 1, as we will show.
    
    \begin{figure*}[!h]    
        \includegraphics[width=1\textwidth]{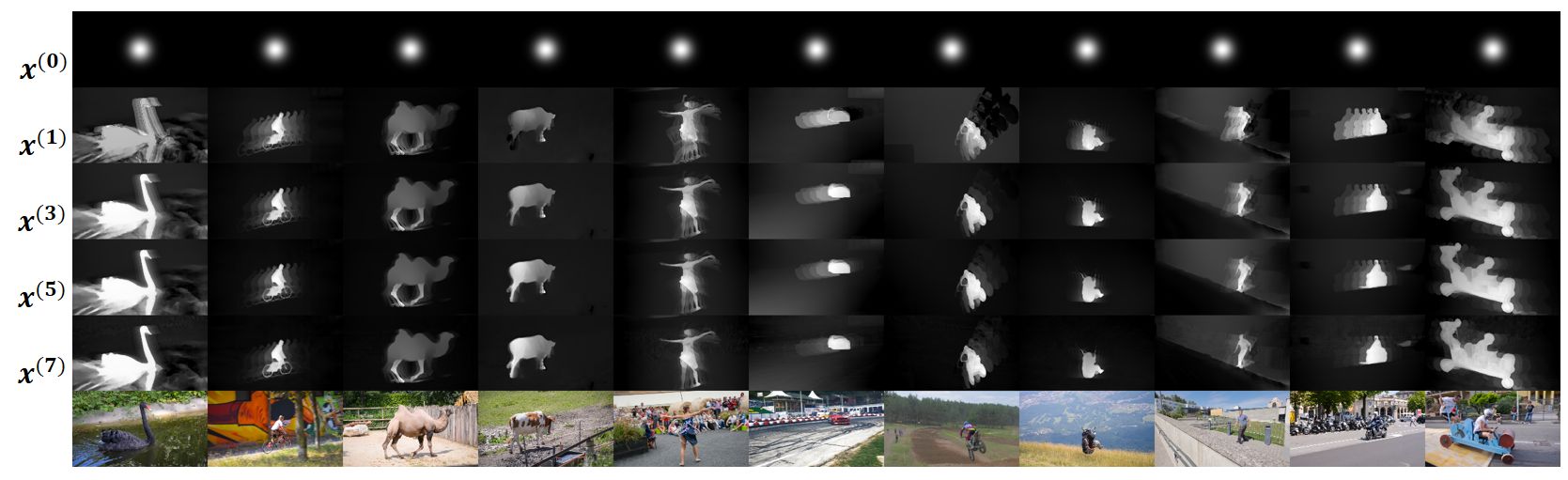}
        \centering
        \caption{Qualitative evolution of the soft masks over 7 iterations of the proposed GO-VOS algorithm, in the fully unsupervised scenario: the segmentation is initialized (first row - $\mathbf{x}^{(0)}$) with a non-informative Gaussian, while the features in $\mathbf{F}$ are only motion directions along optical flow chains. Note how the object of interest is discovered in very few iterations, even though no supervised information or features are used (images from DAVIS2016). }
        \label{fig:first_qualitative_example}
        \vspace{-2.5mm}
    \end{figure*}
    
\section{Algorithm}\label{sec:approach_algorithm_implementation}

    In practice we need to estimate the free parameter $\alpha$, in order to balance in an optimum way the graph term $\mathbf{M}\mathbf{x}$, which depends on node pairs, with the regression term $\mathbf{F}\mathbf{w}$, which depends on features at individual nodes. 
    To keep the algorithm as simple and efficient as possible, we drop $\alpha$ completely and reformulate the iterative process in terms of three separate operations: a propagation step, a regression step and a projection step. This algorithm is equivalent to the power iteration from Eq.~\ref{power_iteration_th} and is guaranteed to converge to the global optimum of Problem 2 (a slightly modified version of Problem 1), as defined in Sec.~\ref{sec:alg_analysis_convergence}.
    
    \textbf{The propagation step} is equivalent to the multiplication $\mathbf{M}\mathbf{x}$, which can be written for a node $i$ as $x_i \gets \sum_j \mathbf{M}_{ij}x_j$. The equation can be implemented efficiently, for all nodes, by propagating the soft labels $x_i$, weighted by the pairwise terms $\mathbf{M}_{ij}$, to all the nodes from the other frames to which $i$ is connected in the graph according to forward and backward optical flow chains. Thus, starting from a given node $i$ we move along the flow chains, one in each direction, and cast node $i$'s votes, $\mathbf{M}_{ij}x_i$, to all points $j$ met along the chain: $x_j \gets x_j + \mathbf{M}_{ij}x_i$. We also increase the value at node $i$ by the same amount: $x_i \gets x_i + \mathbf{M}_{ij}x_j$. By doing so for all pixels in video, in both directions, we perform, in fact, one iteration of $\mathbf{x}^{(it+1)} \gets \mathbf{M}\mathbf{x}^{(it)}$. Since $\mathbf{M}_{ij}=k(i,j)$ decreases rapidly towards zero with the temporal distance between $i$ and $j$ along a chain, in practice we cast votes only between frames that are within a radius of $r$ time steps. That greatly speeds up the process. 
    Thus, the complexity propagation is reduced from $O(n^2)$ to $O(nr)$, where $r$ is a relatively small constant (5 in our experiments). 
    
    \textbf{The regression step} estimates $\mathbf{w}$ for which  $\mathbf{Fw}$ best approximates $\mathbf{x}$ in the least squares error sense. This step is equivalent to ridge regression, where the target values $\mathbf{x}$ are unsupervised, from the solution derived in propagation step.
    
    \textbf{The projection step:} once we compute the optimal $\mathbf{w}$ for current iteration, we can reset the values in $\mathbf{x}$ to be equal to their predicted values $\mathbf{Fw}$. Thus, if the propagation step is a power iteration that pulls the solution towards the main eigenvector of $\mathbf{M}$, the regression and projection steps take the solution closer to the space in which labels can be predicted from actual node features. 
    
    \textbf{Algorithm:} the final GO-VOS algorithm (Alg.~\ref{alg:practical_iterations}) is a slightly simplified version of Eq. \ref{theoretical_iterations} and brings together, in sequence, the three steps discussed above: propagation, regression and projection.
    		
	\begin{algorithm}
        \caption{\textcolor{blue}{GO-VOS}}
        \label{alg:practical_iterations}
        \begin{algorithmic}[1]
            \Statex \textbf{Propagation:}
            \State $\mathbf{x}^{(it+1)} \gets \mathbf{M}\mathbf{x}^{(it)}$ 
            \label{practical_iterations_alg_step1}
            \State $\mathbf{x}^{(it+1)} \gets \mathbf{x}^{(it+1)} / max(\mathbf{x}^{(it+1)})$ \label{practical_iterations_alg_step2}
            \Statex \textbf{Regression:}
            \State $\mathbf{w}^{(it+1)} \gets {(\mathbf{F}^{T}\mathbf{F}+\beta\mathbf{I}_d)}^{-1}\mathbf{F}^{T}\mathbf{x}^{(it+1)}$ \label{practical_iterations_alg_step3}
            \Statex \textbf{Projection:}
            \State $\mathbf{x}^{(it+1)} \gets \mathbf{F}\mathbf{w}^{(it+1)}$ \label{practical_iterations_alg_step4}
        \end{algorithmic}
    \end{algorithm}

    \textbf{Discussion:} in practice it is simpler and also more accurate to compute $\mathbf{w}$ per frame, by using features of nodes from that frame only, such that we get a different $\mathbf{w}$ for each frame. This brings a richer representation power, which explains the superior accuracy.

    \textbf{Initialization:} in the iterative process defined in Alg.~\ref{alg:practical_iterations}, we need to establish the initial labels $\mathbf{x}^{(0)}$ associated to graph nodes. As shown in Sec.~\ref{sec:alg_analysis_init}, irrespective of the initial labels (random labels or informative masks), our algorithm will converge to the same solution, completely defined by motion and features matrices.
    
    The \textbf{adjacency matrix $\mathbf{M}$} is constructed using the optical flow provided by FlowNet2.0~\cite{ilg2017flownet}, which is pretrained on synthetic data (FlyingThings3D \cite{mayer2016large} and FlyingChairs  \cite{dosovitskiy2015flownet}), with no human annotations. In Sec.~\ref{sec:alg_analysis_of} we study the importance of the optical flow quality and also perform tests with the more classical EpicFlow~\cite{revaud2015epicflow}.
    

    \textbf{Features:} the supervision level of our algorithm is mainly dictated by the set of features. 
    The majority of our experiments are performed using \textbf{unsupervised features}, namely colors at the pixel level and motion cues, which are collected along the outgoing optical flow chains as ordered sequences of optical flow displacement between consecutive frames (Figure~\ref{fig:space_time_graph}.b), resulting in a single descriptor $\mathbf{f}_i$. For the case of \textbf{supervised features}, we have only considered foreground probabilities provided by other VOS solutions along the same flow chains and concatenated them to the initial unsupervised $\mathbf{f}_i$. In experiments we show that, as expected, the addition of the stronger features significantly improves performance. Our method can work both as a stand-alone algorithm, able to leverage a wide range of supervised or unsupervised features, as well as a refinement method, by incorporating, in the feature matrix $\mathbf{F}$, the output of other methods.
    
\subsection{Algorithm analysis}\label{sec:approach_algorithm_analysis_and_discussion}
    Further we will perform an in-depth analysis of our algorithm, addressing different aspects, such as convergence, initialization, the set of features and the influence of the optical flow solution. All tests are performed on the validation set of DAVIS2016 \cite{Perazzi2016}, adopting their metrics (Sec.~\ref{sec:experimental_analysis_davis}).
    
    \vspace{-4mm}
    \subsubsection{Convergence to global optimum}\label{sec:alg_analysis_convergence}
    In Sec.~\ref{sec:approach_optimization_problem} we have proved that the power iteration scheme in Eq.~\ref{theoretical_iterations} will converge to the leading eigenvector of a specific matrix $\mathbf{A}$, ensuring that we reach a global optimum. By rewriting the equations of the actual Algorithm~\ref{alg:practical_iterations},  we can demonstrate that our implementation is equivalent to a power iteration scheme, computing the leading eigenvector of another matrix $\mathbf{A}_{alg}$, which we denote the Feature-Motion matrix.
    
    \begin{equation}
        \label{power_iteration}
        \mathbf{x}^{(it+1)} = \frac{\mathbf{A}_{alg}\mathbf{x}^{(it)}}{\|\mathbf{A}_{alg}\mathbf{x}^{(it)}\|_2}, \\
    \end{equation}
    
    where $\mathbf{A}_{alg}=\mathbf{M}\mathbf{F}(\mathbf{F}^{T}\mathbf{F}+\beta\mathbf{I}_d)^{-1}\mathbf{F}^{T}$. Thus, Algorithm ~\ref{alg:practical_iterations} converges to the leading eigenvector of the symmetric Feature-Motion matrix $\mathbf{A}_{alg}$, optimally solving Problem 2, a slightly simplified version of Problem~1 (Eq.~\ref{optimization_problem}):
    \begin{equation}
        \label{optimization_problem_alg}
        \textbf{Problem 2:} \quad \mathbf{x}^{*} = \argmax_{\mathbf{x}} \mathbf{x}^{T}\mathbf{A}_{alg}\mathbf{x} \quad \text{s.t.} \quad \|\mathbf{x}\|_2=1
    \end{equation}

    Note that the only difference between the two segmentation formulations, namely Problem~1 and Problem~2, is the fact that the Feature-Motion matrix $\mathbf{A}_{alg}$ correlates features and motion through a multiplication and provides an exact mathematical description of Algorithm~\ref{alg:practical_iterations}, in contrast to $\mathbf{A}$, expressed as a linear combination of motion and features. The convergence to the principal eigenvector of $\mathbf{A}_{alg}$ implies that the final solution will not depend on the initialization, but only on the graph structure ($\mathbf{M}$) and selected features ($\mathbf{F}$). In Sec.~\ref{sec:alg_analysis_init} we also show that this fact is confirmed by experiments.

    \textbf{Motion structure vs. Feature projection:} The Feature-Motion matrix $\mathbf{A}_{alg}$ can be factored as the product $\mathbf{A}_{alg} = \mathbf{M}\mathbf{P}$, where $\mathbf{M}$ is the motion structure matrix and $\mathbf{P}=\mathbf{F}(\mathbf{F}^{T}\mathbf{F}+\beta\mathbf{I}_d)^{-1}\mathbf{F}^{T}$ is the feature projection matrix.  Thus, at the point of convergence, the segmentation reaches an equilibrium between its motion structure in spacetime and its consistency with the features.
    
    \vspace{-4mm}
    \subsubsection{The role of initialization}\label{sec:alg_analysis_init}
\begin{figure}[h]    
        \includegraphics[width=1\columnwidth]{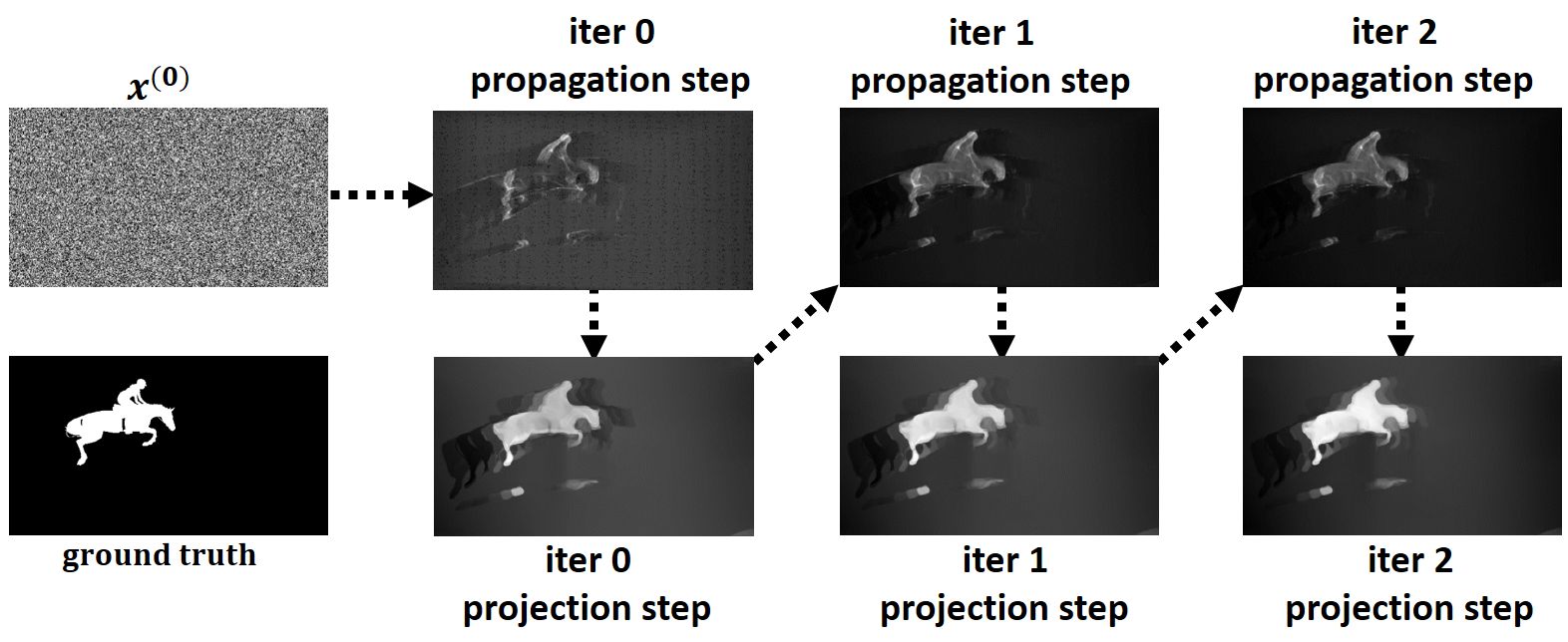}
        \centering
        \caption{Qualitative results of our method, over three iterations, when initialized with a random soft-segmentation mask and using only unsupervised features (i.e. color an motion along flow chains). Note that the main object emerges with each iteration of our algorithm.}
        \label{fig:alg_initialization_figure}
        \vspace{-2.5mm}
    \end{figure}

    Our experiments verify the theoretical results. We observed that the method approaches the same point of convergence, regardless of the initialization. We considered different choices for the initialization $\mathbf{x}^{(0)}$, ranging from uninformative masks such as isotropic Gaussian soft-mask placed in the center of each frame with varied standard deviations, a randomly initialized mask or a uniform full white mask, to masks given by state of the art methods, such as  ELM~\cite{lao2018extending} and PDB~\cite{song2018pyramid}. In Figure \ref{fig:alg_initialization_figure} we present an example, showing the evolution of the soft-segmentation masks over three iterations of our algorithm, when we start from a random mask. We observe that the main object of interest emerges from this initial random mask, as its soft-segmentation mask is visibly improved after each iteration. In Figure~\ref{fig:first_qualitative_example} we present more examples regarding the evolution of our soft-segmentation masks over several iterations. 
    
\begin{figure}[h]    
        \includegraphics[width=1\columnwidth]{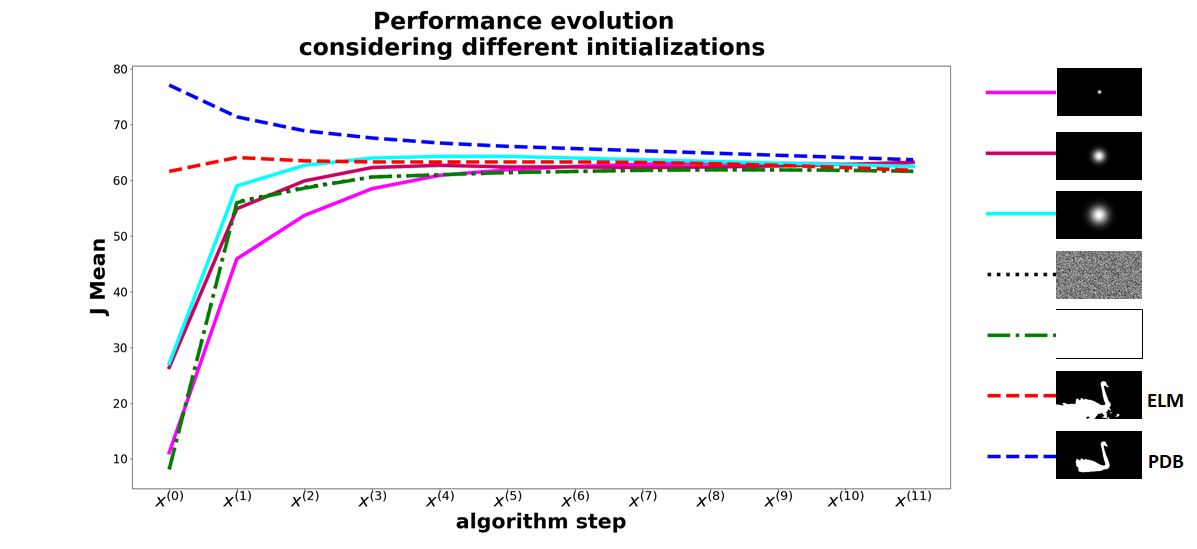}
        \centering
        \caption{Quantitative results of our method considering different initial $\mathbf{x}^{(0)}$, but using the same unsupervised features (color and motion along flow chains). Initializations are (examples given on the right): initial soft-segmentation central Gaussians, with diverse standard deviations, random initial soft mask, uniform white mask and two supervised state of the art solutions PDB~\cite{song2018pyramid} and ELM~\cite{lao2018extending}. Note that, regardless of the initialization, the final metric converges towards the same value, as theoretically proved in Sec. \ref{sec:alg_analysis_convergence}. Tests are performed on full DAVIS validation set.}
        \label{fig:alg_initialization_plot}
         \vspace{-2.5mm}
    \end{figure}

    In Figure~\ref{fig:alg_initialization_plot} we present quantitative results which confirm the theoretical insights. The performance evolves in terms of Jaccard index - J Mean, over seven iterations of our algorithm, towards the same common segmentation. Note that, as expected, convergence is faster for methods that start closer to the convergence point. To conclude, irrespective of the initialization, our implementation will converge towards a unique solution, $\mathbf{x}^*$, that depends only on $\mathbf{M}$ and $\mathbf{F}$, and we have proved this both theoretically and experimentally.

    \vspace{-4mm}
    \subsubsection{The role of features}\label{sec:alg_analysis_features}
    We experimentally study the influence of features, which are expected to have a strong impact on the final result. Our tests clearly show that by adding strong, informative features the performance is boosted significantly. For these experiments we initialized our algorithm with non-informative Gaussian soft-segmentation masks, while considering two sets of features. The first set consists of  motion vectors along the flow chains and will be further referred as the unsupervised set of features. For the second setup, we consider both motion features and foreground probabilities as predicted by the solution of Song \etal \cite{song2018pyramid}(PDB). In the second set we use the prediction of PDB as a strong, targeted supervisory signal, and we will refer to this as the supervised features. Note that even though we have the same starting point, the method converges towards different global optimums, with the supervised set of features producing a more reliable prediction.

    \begin{figure}[h]    
        \includegraphics[width=1\columnwidth]{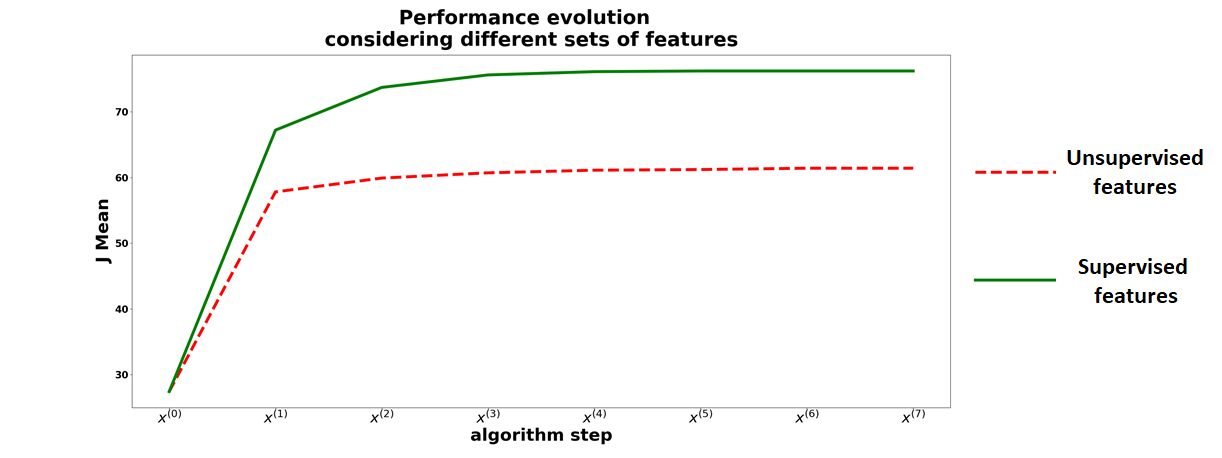}
        \centering
        \caption{Performance evolution over seven iterations, considering different sets of features. Tests are performed on full DAVIS validation set.}
        \label{fig:features_plot}
        \vspace{-2.5mm}
    \end{figure}
    
    \vspace{-4mm}
    \subsubsection{The role of optical flow}\label{sec:alg_analysis_of}
    The convergence point of our solution is completely defined by motion matrix $\mathbf{M}$ and features matrix $\mathbf{F}$. Considering that the motion matrix is constructed using the optical flow, the quality of our solution will be influenced by the quality of the optical flow solution. We have used FlowNet2.0 \cite{ilg2017flownet} for our algorithm, but here we consider replacing FlowNet2.0 with EpicFlow \cite{revaud2015epicflow}. In Table~\ref{tab:experiments_optical_flow} we present quantitative results of our experiments, comparing our solution against other unsupervised solutions published on DAVIS2016 (for details we refer the reader to Sec.~\ref{sec:experimental_analysis_davis}). EpicFlow is less accurate than FlowNet2.0 and this is also reflected in performance of our solution, but we are still having competitive results, being on second place among the considered solutions. 
    \begin{table}[h]
    \begin{center}
    \resizebox{0.95\columnwidth}{!}{%
    \begin{tabular}{|c|c|c|c||c|c|}
        \hline
        ELM[15] & FST[25] & CUT[13] & NLC[8] & \textbf{\textcolor{blue}{GO-VOS}+EpicFlow} & \textbf{\textcolor{blue}{GO-VOS}+FlowNet2.0} \\
        \hline
        61.8 & 55.8 & 55.2 & 55.1 & 61.0 & \textbf{\textcolor{blue}{65.0}} \\
        \hline
    \end{tabular}
    }
    \end{center}
    \caption{Quantitative results of our method, compared with state of the art solutions on DAVIS2016, considering different optical flow solutions.}
    \label{tab:experiments_optical_flow}
    \vspace{-6mm}
    \end{table}

\section{Experiments}\label{sec:experimental_analysis}
\begin{figure*}[h]   
        \centering
        \includegraphics[width=1\textwidth]{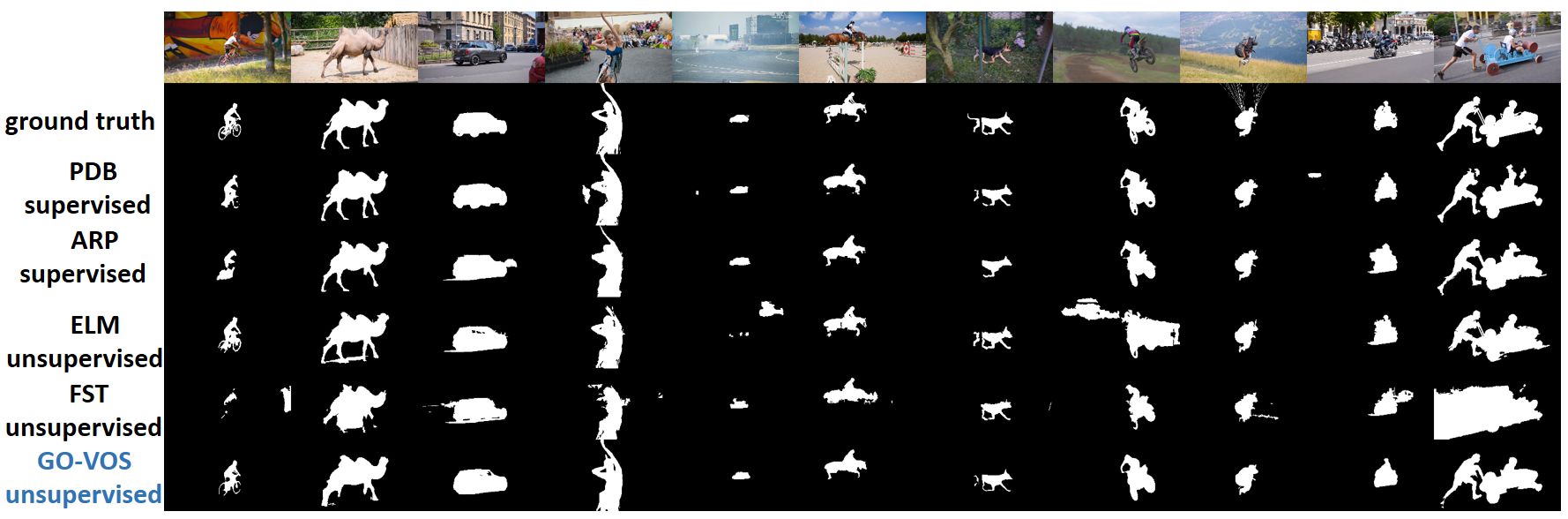}
        \caption{Qualitative results of the proposed GO-VOS algorithm, in the fully unsupervised case. We also present results of four other approaches: PDB \cite{song2018pyramid}, ARP\cite{koh2017primary}, ELM \cite{lao2018extending} and FST \cite{papazoglou2013fast}.}
        \label{fig:qualitative_results}
    \end{figure*}

    We compare our proposed approach, GO-VOS, against state of the art solutions for video object segmentation, on three challenging datasets DAVIS \cite{Perazzi2016}, SegTrack v2 \cite{li2013video} and YouTube-Objects \cite{kalogeiton2016analysing}. In all the experiments we initialized the soft-segmentation masks with non-informative Gaussian soft-masks placed in the center of each frame. Unless otherwise specified, we used only unsupervised features: motion and color cues. We present some qualitative results in Figure~\ref{fig:qualitative_results}, in comparison to other methods on the DAVIS2016 dataset, which we present next.
    
\subsection{DAVIS dataset}\label{sec:experimental_analysis_davis}
Perazzi \etal introduce in \cite{Perazzi2016} a dataset and an evaluation methodology for VOS tasks. The original dataset is composed of 50 videos, each accompanied by accurate, per pixel annotations of the main object of interest. For the initial version of the dataset the main object can be composed of multiple strongly connected objects, considered as a single object. DAVIS is a challenging dataset as it contains many difficult cases such as appearance changes, occlusions and motion blur. \textbf{Metric:} For evaluation we compute both region-based (J Mean) and contour-based (F Mean) measures as established in \cite{Perazzi2016}. J Mean is computed as the intersection over union between the estimated segmentation and the ground truth. F Mean is the F-measure of the segmentation contour points (for details see \cite{Perazzi2016}). \textbf{Results:} In Table~\ref{tab:experiments_davis_results} we compare our method (GO-VOS) against both supervised and unsupervised methods on the task of unsupervised single object segmentation on DAVIS validation set (20 videos). To compare GO-VOS against a supervised solution, we include the predictions of that method in our features matrix, while initializing our solution with a non-informative mask. This can be seen as a refinement step. We highlight that our unsupervised GO-VOS achieves state of the art results among unsupervised methods and also improves over the ones that use supervised pretrained features.
Note that a method is considered unsupervised if it requires no training using human annotated object segmentation.

\begin{table}[h]
    \begin{center}
    \resizebox{0.95\columnwidth}{!}{%
    \begin{tabular}{|c|c|l|c|c|c|}
        \hline
        Task & \multicolumn{2}{c|}{Method} & J Mean & F Mean & sec/frame \\
        \hline\hline
         \multirow{16}{*}{Unsupervised} 
            & \multirow{10}{*}{\shortstack[]{Supervised \\ features}}
             & PDB\cite{song2018pyramid} & 77.2 & 74.5 & \textbf{0.05} \\\cline{3-6}
            & & ARP\cite{koh2017primary} & 76.2 & 70.6 & N/A \\\cline{3-6}
            & & LVO\cite{tokmakov2017learning} & 75.9 & 72.1 & N/A \\\cline{3-6}
            & & FSEG\cite{jain2017fusionseg} & 70.7 & 65.3 & N/A \\\cline{3-6}
            & & LMP\cite{tokmakov2017learning_} & 70.0 & 65.9 & N/A 
            \\\hhline{~~====}
            & & \shortstack[l]{\textbf{\textcolor{blue}{GO-VOS supervised}} \\ + features of \cite{song2018pyramid}} & \textbf{79.9} (\textcolor{green}{+2.7}) & \textbf{78.1} & 0.61 \\\cline{3-6}
            &  & \shortstack[l]{\textbf{\textcolor{blue}{GO-VOS supervised}} \\ + features of \cite{koh2017primary}} & 78.7 (\textcolor{green}{+2.5}) & 73.1 & 0.61\\\cline{3-6}
            & & \shortstack[l]{\textbf{\textcolor{blue}{GO-VOS supervised}}\\+ features of \cite{tokmakov2017learning}} & 77.0 (\textcolor{green}{+1.1}) & 73.7 & 0.61\\\cline{3-6}
            & & \shortstack[l]{\textbf{\textcolor{blue}{GO-VOS supervised}}\\+ features of \cite{jain2017fusionseg}} & 74.1 (\textcolor{green}{+3.5}) & 69.9 & 0.61 \\\cline{3-6}
            &  & \shortstack[l]{\textbf{\textcolor{blue}{GO-VOS supervised}}\\+ features of \cite{tokmakov2017learning_}} & 73.7 (\textcolor{green}{+3.7})& 69.2 & 0.61
            \\\hhline{~=====}
            & \multirow{6}{*}{Unsupervised} 
              & ELM\cite{lao2018extending} & 61.8 & \textbf{\textcolor{blue}{61.2}} & 20 \\\cline{3-6}
             & & FST\cite{papazoglou2013fast} & 55.8 & 51.1 & 4\\\cline{3-6}
             & & CUT\cite{keuper2015motion} & 55.2 & 55.2 & $\approx$1.7 \\\cline{3-6} 
             & & NLC\cite{faktor2014video} & 55.1 & 52.3 & 12 \\\cline{3-6} 
             & & \textbf{\textcolor{blue}{GO-VOS unsupervised}} & \textbf{\textcolor{blue}{65.0}} & 61.1 & \textbf{\textcolor{blue}{0.61}} \\
    \hline
    \end{tabular}
    }
    \end{center}
    \caption{Quantitative results of our method, compared with state of the art solutions on DAVIS2016. Our proposed method achieves top results, while also being the fastest among unsupervised methods. \cite{song2018pyramid} is much faster as it is based on a trained network, requiring only a forward pass at test time. (black bold - best supervised; blue bold - best unsupervised)}
    \label{tab:experiments_davis_results}
     \vspace{-6mm}
\end{table}

\subsection{SegTrack v2 dataset}\label{sec:experimental_analysis_segtrack}
The SegTrack dataset was originally introduced in \cite{tsai2012motion} and further adapted for the VOS task in \cite{li2013video}. SegTrack v2 contains 14 videos with pixel level annotations for the main objects of interest (8 videos with a single object and 6 containing multiple objects). SegTrack contains deformable and dynamic objects, with videos at a relative poor resolution - making it a very challenging dataset for video object segmentation. \textbf{Metric:} For evaluation we used the average intersection over union score. \textbf{Results:} In Table~\ref{tab:experiments_segtrack_results} we present quantitative results of our method, using only unsupervised features. Our solution is surpassed by NLC \cite{faktor2014video}, but we highlight that we have better results than NLC for DAVIS2016. In terms of speed, GO-VOS is the second best. 

\begin{table}[h]
    \begin{center}
    \resizebox{0.85\columnwidth}{!}{%
    \begin{tabular}{|c|c|l|c|c|}
        \hline
        Task & \multicolumn{2}{c|}{Method} & IoU & sec/frame \\
        \hline\hline
        \multirow{8}{*}{Unsupervised} 
            & \multirow{4}{*}{\shortstack[]{Supervised \\features}}
             & KEY \cite{lee2011key} & 57.3 & $>$120 \\\cline{3-5}
            & & FSEG \cite{jain2017fusionseg} & \textbf{61.4} & N/A \\\cline{3-5}
            & & LVO \cite{tokmakov2017learning} & 57.3 & N/A \\\cline{3-5}
            & & \cite{li2018instance} & 59.3 & N/A
            \\\hhline{~====}
            & \multirow{4}{*}{Unsupervised} 
             & NLC \cite{faktor2014video} & \textbf{\textcolor{blue}{67.2}} & 12 \\\cline{3-5}
             & & FST \cite{papazoglou2013fast} & 54.3 & 4 \\\cline{3-5}
             & & CUT \cite{keuper2015motion} & 47.8 & $\approx$1.7 \\\cline{3-5}
             & & HPP \cite{haller2017unsupervised} & 50.1 & \textbf{\textcolor{blue}{ 0.35}}
             \\\hhline{~~===}
             & & \textbf{\textcolor{blue}{GO-VOS unsupervised}} & 62.2 & 0.61 \\
    \hline
    \end{tabular}
    }
    \end{center}
    \caption{Quantitative results of our method, compared with state of the art solutions on SegTrack v2 dataset. (black bold - best supervised; blue bold - best unsupervised)}
    \label{tab:experiments_segtrack_results}
    \vspace{-6mm}
\end{table}

\subsection{YouTube-Objects dataset}\label{sec:experimental_analysis_yto}
YouTube-Objects (YTO) dataset \cite{prest2012learning} consists of videos collected from YouTube ($\approx$ 720k frames). It is very challenging, containing 2511 video shots, with the ground truth provided in the form of object bounding boxes. Although there are no pixel level annotations, YTO tests are relevant considering the large number of videos and wide diversity. In the paper, we present results on v2.2, containing more annotated boxes (6975), but we also provide state of the art results on v1.0 in the supplementary material. Following the methodology of published works, we test our solution on the train set, which contains videos with only one annotated object. \textbf{Metric:} We used the CorLoc metric, computing the percentage of correctly localized object bounding boxes, according to PASCAL-criterion (IoU $\geq 0.5$). \textbf{Results:} In Table~\ref{tab:experiments_yto_results} we present the results on YTO v2.2 and compare against the published state of the art. All methods are fully unsupervised. We obtain the top average score, while outperforming the other methods on 5 out of 10 object classes.

    \begin{table}[h]
        \begin{center}
        \resizebox{\columnwidth}{!}{%
        \begin{tabular}{|c||c|c|c|c|c|c|c|c|c|c||c||c|}
            \hline
            Method & aero & bird & boat & car & cat & cow & dog & horse & moto & train & avg & sec/frame\\
            \hline\hline
            \cite{croitoru2017unsupervised} & 75.7 & 56.0 & 52.7 & 57.3 & 46.9 & \textbf{\textcolor{blue}{57.0}} & 48.9 & 44.0 & 27.2 & 56.2 & 52.2 & \textbf{\textcolor{blue}{0.02}} \\
            \hline
            HPP\cite{haller2017unsupervised} & 76.3 & 68.5 & \textbf{\textcolor{blue}{54.5}} & 50.4 & \textbf{\textcolor{blue}{59.8}} & 42.4 & 53.5 & 30.0 & \textbf{\textcolor{blue}{53.5}} & \textbf{\textcolor{blue}{60.7}} & 54.9 & 0.35 \\
            \hline\hline
            \shortstack[]{\textcolor{blue}{\textbf{GO-VOS}} \\ \textcolor{blue}{\textbf{unsupervised}}} & \textbf{\textcolor{blue}{79.8}} & \textbf{\textcolor{blue}{73.5}} & 38.9 & \textbf{\textcolor{blue}{69.6}} & 54.9 & 53.6 & \textbf{\textcolor{blue}{56.6}} & \textbf{\textcolor{blue}{45.6}} & 52.2 & 56.2 & \textbf{\textcolor{blue}{58.1}} & 0.61 \\
            \hline
        \end{tabular}
        }
        \end{center}
        \caption{Quantitative results of our method (CorLoc metric), compared with state of the art solutions on YTO v2.2. We have top results for 5 out of 10 classes, and in average. \cite{croitoru2017unsupervised} is faster as it is based on a trained network, requiring only a forward pass at test time.  (blue bold - best solution)}
        \label{tab:experiments_yto_results}
        \vspace{-6mm}
    \end{table}
    
\subsection{Computation cost}\label{sec:computation_time}
    Considering our formulation, as a graph with a node per each video pixel and long range connections, one would expect it to be memory expensive and slow. However, since we do not actually construct the adjacency matrix and have an efficient implementation for our algorithm steps, complexity is only $O(n)$ ($n$ is the number of video pixels). We require 0.04 sec/frame for computing the optical flow and 0.17 sec/frame for computing information related to matrices $\mathbf{M}$ and $\mathbf{F}$. Further, our optimization process requires 0.4 sec/frame, resulting in a total of 0.61 sec/frame for the full algorithm. Our solution is implemented in PyTorch and the runtime analysis is performed on a computer with specifications: Intel(R) Xeon(R) CPU E5-2697A v4 @ 2.60GHz, GPU GeForce GTX 1080. 

\section{Conclusions}\label{sec:conclusions}
    We present a novel graph representation at the dense pixel level for the problem of foreground object segmentation. We provide a spectral clustering formulation, in which the optimal solution is computed fast, by power iteration, as the principal eigenvector of a novel Feature-Motion matrix. The matrix couples local information at the level of pixels as well as long range connections between pixels through optical flow chains. In this view, objects become strong, principal clusters of motion and appearance patterns in their immediate spacetime neighborhood. Thus, the two "forces" in space and time, expressed through motion and appearance, are brought together into a single power iteration formulation that reaches the global optimum in a few steps. In extensive experiments, we show that the proposed algorithm, GO-VOS, is fast and obtains state of the art results on three challenging benchmarks used in current literature.
    
{\small
\bibliographystyle{ieee}
\bibliography{bib}
}
\end{document}


\title{Supplementary material \\ Spacetime Graph Optimization for Video Object Segmentation}

\author{First Author \\
Institution1\\
{\tt\small firstauthor@i1.org}
\and
Second Author \\
Institution2\\
{\tt\small secondauthor@i2.org}
}

\maketitle
\ifwacvfinal\thispagestyle{empty}\fi

First two sections of this material provide theoretical proofs of the properties of our algorithm, while the third section introduces a complexity analysis of our algorithm. The fourth and fifth sections provide additional quantitative and qualitative results.

In Section~\ref{convergence_proof} we provide an in depth analysis regarding the convergence of the algorithm introduced in our original paper, and reiterated in Algorithm~\ref{alg:practical_iterations}. Additionally, we theoretically prove that the proposed algorithm optimally solves our objective: Problem 2 (Eq.~\ref{optimization_problem_alg}). Section~\ref{video_level_frame_level} addresses the convergence of the algorithm in the setup where $\mathbf{w}$ is learned per frame, proving that the resulted algorithm (Algorithm~\ref{alg:practical_iterations_per_frame}) is equivalent to Algorithm~\ref{alg:practical_iterations}, where we replace features matrix $\mathbf{F}$ with the block features matrix $\mathbf{F}_f$. Algorithm~\ref{alg:practical_iterations_per_frame} has the same properties of as Algorithm~\ref{alg:practical_iterations} and will converge to the leading eigenvector of matrix $\mathbf{A}_{alg}^{f}$, optimally solving Problem 3. We highlight that $\mathbf{A}_{alg}^{f}$ is the same as $\mathbf{A}_{alg}$, where we have replaced $\mathbf{F}$ with block matrix $\mathbf{F}_f$.

Section~\ref{complexity} presents a detailed analysis of the complexity of the proposed algorithm, for a better understanding of its efficiency. 

Section~\ref{quantitative_res} introduces state of the art quantitative results for v1.0 of YouTube-Objects(YTO) dataset, as stated in our original paper, where we have presented only the results for v2.2 of YTO. In Section~\ref{qualitative_res} we present few qualitative results on YTO and SegTrack datasets. 

\section{Convergence analysis}\label{convergence_proof}
    The proposed algorithm, GO-VOS, solves the following problem, depicted in our paper as Problem 2:
    \begin{equation}
        \label{optimization_problem_alg}
        \begin{aligned}
        \textbf{Problem 2:} \quad \mathbf{x}^{*} = \argmax_{\mathbf{x}} \mathbf{x}^{T}\mathbf{A}_{alg}\mathbf{x} \quad \text{s.t.} \quad \|\mathbf{x}\|_2=1 \\
        \text{where }
        \mathbf{A}_{alg}=\mathbf{M}\mathbf{F}(\mathbf{F}^{T}\mathbf{F}+\beta\mathbf{I}_d)^{-1}\mathbf{F}^{T}
        \end{aligned}
    \end{equation}
    
    We will further prove that Algorithm~\ref{alg:practical_iterations} converges to the leading eigenvector of Feature-Motion matrix $\mathbf{A}_{alg}$ (Sec.~\ref{prop_1} - Proposition 1) and that this is a global optimum for our problem defined in Eq.~\ref{optimization_problem_alg} (Sec.~\ref{prop_2} - Proposition 2). 
    
    \begin{algorithm}
        \caption{\textcolor{blue}{GO-VOS}}
        \label{alg:practical_iterations}
        \begin{algorithmic}[1]
            \Statex \textbf{Propagation:}
            \State $\mathbf{x}^{(it+1)} \gets \mathbf{M}\mathbf{x}^{(it)}$ 
            \label{practical_iterations_alg_step1}
            \State $\mathbf{x}^{(it+1)} \gets \mathbf{x}^{(it+1)} / max(\mathbf{x}^{(it+1)})$ \label{practical_iterations_alg_step2}
            \Statex \textbf{Regression:}
            \State $\mathbf{w}^{(it+1)} \gets {(\mathbf{F}^{T}\mathbf{F}+\beta\mathbf{I}_d)}^{-1}\mathbf{F}^{T}\mathbf{x}^{(it+1)}$ \label{practical_iterations_alg_step3}
            \Statex \textbf{Projection:}
            \State $\mathbf{x}^{(it+1)} \gets \mathbf{F}\mathbf{w}^{(it+1)}$ \label{practical_iterations_alg_step4}
        \end{algorithmic}
    \end{algorithm}
    
    \subsection{Convergence analysis of Algorithm 1}\label{prop_1}
    \textbf{Proposition 1:} Algorithm 1 converges to the leading eigenvector of Feature-Motion matrix $\mathbf{A}_{alg}$.
    
    \textbf{Proof:} Starting from Algorithm 1, we will express $x^{(it+1)}$ only function of  matrices $\mathbf{M}$ and $\mathbf{F}$, and its previous value $\mathbf{x}^{it}$. 
    
    Notations:
    \begin{itemize}
        \item ${\text{Step}_i}^{(it)}$ - Step $i$ of Algorithm~\ref{alg:practical_iterations}, at iteration $it$
    \end{itemize}
    
    For our algorithm, only the relative order of node labels is relevant, and the max-normalization of $\text{Step}_2^{(it+1)}$ keeps the direction of $\mathbf{x}$ unchanged, being simple and efficient. Considering that the direction of $\mathbf{x}$ remains unchanged, in the context of our algorithm, it is equivalent with applying a L2-normalization: $\mathbf{x}^{(it+1)}=\frac{\mathbf{x}^{(it+1)}}{\|\mathbf{x}^{(it+1)}\|_2}$, which we will further consider as $\text{Step}_2^{(it+1)}$.
    
    \begin{equation}
        \label{eq_merge_1_and_2_it_plus_1}
        \begin{rcases}
             {\text{Step}_1}^{(it+1)}\text{: }\mathbf{x}^{(it+1)} \gets \mathbf{M}\mathbf{x}^{(it)}\\
             {\text{Step}_2^{(it+1)}\text{: }\mathbf{x}^{(it+1)} \gets \frac{\mathbf{x}^{(it+1)}}{\|\mathbf{x}^{(it+1)}\|_2}}
        \end{rcases} \Rightarrow \mathbf{x}^{(it+1)}=\frac{\mathbf{M}\mathbf{x}^{(it)}}{\|\mathbf{M}\mathbf{x}^{(it)}\|_2}
    \end{equation}
    
    \begin{equation}
        \label{eq_merge_4}
        \begin{rcases}
            \text{Eq.~\ref{eq_merge_1_and_2_it_plus_1}: }\mathbf{x}^{(it+1)}=\frac{\mathbf{M}\mathbf{x}^{(it)}}{\|\mathbf{M}\mathbf{x}^{(it)}\|_2} \\ 
             {\text{Step}_4}^{(it)}\text{ : }\mathbf{x}^{(it)} \gets \mathbf{F}\mathbf{w}^{(it)}
        \end{rcases} \Rightarrow \mathbf{x}^{(it+1)} = \frac{\mathbf{M}\mathbf{F}\mathbf{w}^{(it)}}{\|\mathbf{M}\mathbf{F}\mathbf{w}^{(it)}\|_2}
    \end{equation}
    
    \begin{equation}
        \label{eq_merge_3}
        \begin{aligned}
        \begin{rcases}
             \text{Eq.~\ref{eq_merge_4}: }\mathbf{x}^{(it+1)} = \frac{\mathbf{M}\mathbf{F}\mathbf{w}^{(it)}}{\|\mathbf{M}\mathbf{F}\mathbf{w}^{(it)}\|_2}\\
             {\text{Step}_3}^{(it)}\text{ : }\mathbf{w}^{(it)} \gets
             (\mathbf{F}^{T}\mathbf{F}+\beta\mathbf{I}_d)^{-1}\mathbf{F}^{T}\mathbf{x}^{(it)}
        \end{rcases} \Rightarrow \\
        \Rightarrow
        \mathbf{x}^{(it+1)}=\frac{1}{p}\mathbf{M}\mathbf{F}(\mathbf{F}^{T}\mathbf{F}+\beta\mathbf{I}_d)^{-1}\mathbf{F}^{T}\mathbf{x}^{(it)} \\
        \text{where }p=\|\mathbf{M}\mathbf{F}(\mathbf{F}^{T}\mathbf{F}+\beta\mathbf{I}_d)^{-1}\mathbf{F}^{T}\mathbf{x}^{(it)}\|_2
        \end{aligned}
    \end{equation}
    
    Further, Eq.~\ref{eq_merge_3} can be written as:
    
    \begin{equation}
        \label{eq_final}
        \mathbf{x}^{(it+1)} = \frac{\mathbf{A}_{alg}\mathbf{x}^{(it)}}{\|\mathbf{A}_{alg}\mathbf{x}^{(it)}\|_2} \\
    \end{equation}
     
    where $\mathbf{A}_{alg}=\mathbf{M}\mathbf{F}(\mathbf{F}^{T}\mathbf{F}+\beta\mathbf{I}_d)^{-1}\mathbf{F}^{T}$. 
    
    \hfill \quad $\square$
    
    We have proved that Algorithm~\ref{alg:practical_iterations} converges to the leading eigenvector of Feature-Motion matrix $\mathbf{A}_{alg}$ (Eq.~\ref{eq_final}). Although it is well known that Eq.~\ref{eq_final} solves Problem 2, we will further provide a formal proof. 
    
    \subsection{Convergence point analysis}\label{prop_2}
    \textbf{Proposition 2:} The leading eigenvector of matrix $\mathbf{A}_{alg}$ is the global optimum of Problem 2.
    
    \textbf{Proof:} In Problem 2, the function $\mathbf{x}^{T}\mathbf{A}_{alg}\mathbf{x}$ should be maximized under the equality constraint $\mathbf{x}^{T}\mathbf{x}=1$. Following the proof of Problem 1 from our original paper, we introduce the Lagrange multiplier $\lambda_{alg}$ and define the Lagrange function:
    
    \begin{equation}
        \label{alg_lagrange_fct}
        \mathcal{L}_{alg}(\mathbf{x}, \lambda_{alg})=\mathbf{x}^{T}\mathbf{A}_{alg}\mathbf{x}-\lambda_{alg}(\mathbf{x}^{T}\mathbf{x}-1)
    \end{equation}
   
    The solution is searched among the stationary points satisfying $\nabla_{\mathbf{x},\lambda_{alg}}\mathcal{L}_{alg}(\mathbf{x},\lambda_{alg})=\mathbf{0}$. We prove that we arrive at a stationary point that is also the global optimum, the leading eigenvector of Feature-Motion matrix $\mathbf{A}_{alg}$.
    
    \begin{equation}
        \label{alg_lagrange_deriv}
        \begin{cases}
            \nabla_{\mathbf{x}}\mathcal{L}_{alg}(\mathbf{x},\lambda_{alg})=\mathbf{0} \Rightarrow \mathbf{A}_{alg}\mathbf{x}-\lambda_{alg}\mathbf{x}=\mathbf{0} \\
            \nabla_{\lambda_{alg}}\mathcal{L}_{alg}(\mathbf{x}, \lambda_{alg})=0 \Rightarrow \mathbf{x}^{T}\mathbf{x}=1
        \end{cases}
    \end{equation}
    
    We will rewrite Eq.~\ref{alg_lagrange_deriv} in the form $\mathbf{x}=h_{alg}(\mathbf{x})$ and any fixed point of $h_{alg}$ will be a solution for our initial problem. Similar to Problem 1, we apply a fixed point iteration scheme and iteratively update $\mathbf{x}$ function of its previous value. The function $h_{alg}$ is defined as: $h_{alg}(\mathbf{x})=\frac{1}{p'_{alg}\lambda_{alg}}\mathbf{A}_{alg}\mathbf{x}$, where $p'_{alg}=\|\frac{1}{\lambda_{alg}}\mathbf{A}_{alg}\mathbf{x}\|_2$. We observe that the term $\lambda_{alg}$ cancels out, resulting our final iterative formulation:
    
    \begin{equation}
        \label{alg_iterative_form}
        \begin{aligned}
        \mathbf{x}^{(it+1)}=\frac{1}{p_{alg}}\mathbf{A}_{alg}\mathbf{x}^{(it)} \\
        \text{where }p_{alg}=\|\mathbf{A}_{alg}\mathbf{x}^{(it)}\|_2
        \end{aligned}
    \end{equation}

    According to Eq.~\ref{alg_iterative_form}, the leading eigenvector of $\mathbf{A}_{alg}$ is indeed the global optimum of Problem 2. 
    
    \hfill \quad $\square$
    
    This proves that the convergence point of Algorithm~\ref{alg:practical_iterations} is the global optimum of Problem 2. 

\section{Convergence analysis when learning $\mathbf{w}$ per frame}\label{video_level_frame_level}
    In practice, it can be more simpler to compute $\mathbf{w}$ per frame, instead of per video, which we have also specified in our original paper. When dealing with low level features, this should also be more accurate. Further, we will prove that even if we consider computing $\mathbf{w}$ per frame, our algorithm is still guaranteed to converge to the leading eigenvector of a different Feature-Motion matrix $\mathbf{A}_{alg}^{f}$, which differs from $\mathbf{A}_{alg}$ in the definition of the considered features matrix. 
    
    Notations: 
    \begin{itemize}
        \item $\mathbf{F}_t \in \mathbb{R}^{hw \times d}$ - feature matrix associated to frame $t$
        \item $\mathbf{x}_t \in \mathbb{R}^{hw \times 1}$ - labels of nodes associated to frame $t$
         $\mathbf{x}=$ \begin{pmatrix}
                \mathbf{x}_0 \\
                \mathbf{x}_1\\
                \vdots \\
                \mathbf{x}_m 
        \end{pmatrix} $\in \mathbb{R}^{n \times 1}$
        \item $\mathbf{w}_t \in \mathbb{R}^{d \times 1}$ - model for frame $t$
        \item $\mathbf{F}_{f}=$
         \begin{pmatrix}
                \mathbf{F}_0 & \mathbf{0} & \dots & \mathbf{0} \\
                \mathbf{0} & \mathbf{F}_1 & \dots & \mathbf{0} \\
                \vdots & \vdots & \ddots & \vdots \\
                \mathbf{0} & \mathbf{0} & \dots & \mathbf{F}_m
        \end{pmatrix} $\in \mathbb{R}^{n \times dm}$
        \item $\mathbf{w}_{f}=$
         \begin{pmatrix}
                \mathbf{w}_0 \\
                \mathbf{w}_1\\
                \vdots \\
                \mathbf{w}_m 
        \end{pmatrix} $\in \mathbb{R}^{dm \times 1}$
    \end{itemize}
    
    where $m$ is the number of frames, $(h,w)$ the size of a frame and $d$ is the number of features per node. 
    
    \textbf{Proposition 3:} Computing $\mathbf{w}$ per frame is equivalent to replacing $\mathbf{F}$ with $\mathbf{F}_f$ in Algorithm~\ref{alg:practical_iterations}.
    
    \textbf{Proof:} 
    According to $\text{Step}_4^{(it+1)}$, $\mathbf{x}^{(it+1)}=\mathbf{F}\mathbf{w}^{(it+1)}$ and if we assume frame level regression, for a given frame $t$, $\mathbf{x}_t^{(it+1)}=\mathbf{F}_t\mathbf{w}_t^{(it+1)}$. It results that we can express $\mathbf{x}^{(it+1)}$ as follows:
    
    \begin{equation}
        \label{equiv_step_4}
        \mathbf{x}^{(it+1)}=\begin{pmatrix}\mathbf{x}_0^{(it+1)}\\ \mathbf{x}_1^{(it+1)}\\
    \vdots \\ \mathbf{x}_m^{(it+1)}\end{pmatrix}=\begin{pmatrix}\mathbf{F}_0\mathbf{w}_0^{(it+1)} \\ \mathbf{F}_1\mathbf{w}_1^{(it+1)}\\
    \vdots \\
    \mathbf{F}_m\mathbf{w}_m^{(it+1)}\end{pmatrix}=\mathbf{F}_f\mathbf{w}_f^{(it+1)}
    \end{equation}
    
    $\text{Step}_3^{(it+1)}$ computes $\mathbf{w}^{(it+1)}$ as $(\mathbf{F}^{T}\mathbf{F}+\beta\mathbf{I}_d)^{-1}\mathbf{F}^{T}\mathbf{x}^{(it+1)}$ and assuming frame level regression $\mathbf{w}_t^{(it+1)}=(\mathbf{F}_t^{T}\mathbf{F}_t+\beta\mathbf{I}_d)^{-1}\mathbf{F}_t^{T}\mathbf{x}_t^{(it+1)}$. It results that we can express $\mathbf{w}^{(it+1)}$ as follows:
    
    $\mathbf{w}^{(it+1)}=\begin{pmatrix}\mathbf{w}_0^{(it+1)}\\ \mathbf{w}_1^{(it+1)} \\ \vdots \\ \mathbf{w}_m^{(it+1)}\end{pmatrix}= \\ =\begin{pmatrix}
    (\mathbf{F}_0^{T}\mathbf{F}_0+\beta\mathbf{I}_d)^{-1}\mathbf{F}_0^{T}\mathbf{x}_0^{(it+1)} \\
    (\mathbf{F}_1^{T}\mathbf{F}_1+\beta\mathbf{I}_d)^{-1}\mathbf{F}_1^{T}\mathbf{x}_1^{(it+1)} \\
    \vdots \\
    (\mathbf{F}_m^{T}\mathbf{F}_m+\beta\mathbf{I}_d)^{-1}\mathbf{F}_m^{T}\mathbf{x}_m^{(it+1)}
    \end{pmatrix}= \\
    =\begin{pmatrix}
    \mathbf{R}_0^{-1} & \mathbf{0} & \dots & \mathbf{0} \\
    \mathbf{0} & \mathbf{R}_1^{-1} & \cdots & \mathbf{0} \\
     \vdots & \vdots & \ddots & \vdots \\
    \mathbf{0} & \mathbf{0} & \dots & 
    \mathbf{R}_m^{-1}
    \end{pmatrix}\cdot \begin{pmatrix}
    \mathbf{F}_0^{T} & \mathbf{0} & \dots & \mathbf{0} \\
    \mathbf{0} & \mathbf{F}_1^{T} & \cdots & \mathbf{0} \\
     \vdots & \vdots & \ddots & \vdots \\
    \mathbf{0} & \mathbf{0} & \dots & 
    \mathbf{F}_m^{T}
    \end{pmatrix}\cdot$
    
    \cdot\begin{pmatrix}\mathbf{x}_0^{(it+1)} \\ \mathbf{x}_1^{(it+1)} \\ \vdots \\ \mathbf{x}_m^{(it+1)}\end{pmatrix}   \text{, where }$\mathbf{R}_t = \mathbf{F}_t^{T}\mathbf{F}_t+\beta\mathbf{I}_d$.
    
    $\Rightarrow \mathbf{w}^{(it+1)}=\begin{pmatrix}
    \mathbf{R}_0^{-1} & \mathbf{0} & \dots & \mathbf{0} \\
    \mathbf{0} & \mathbf{R}_1^{-1} & \cdots & \mathbf{0} \\
     \vdots & \vdots & \ddots & \vdots \\
    \mathbf{0} & \mathbf{0} & \dots & 
    \mathbf{R}_m^{-1}
    \end{pmatrix}\mathbf{F}_f^{T}\mathbf{x}^{(it+1)}=$
    
    $=\begin{pmatrix}
    \mathbf{R}_0 & \mathbf{0} & \dots & \mathbf{0} \\
    \mathbf{0} & \mathbf{R}_1 & \cdots & \mathbf{0} \\
     \vdots & \vdots & \ddots & \vdots \\
    \mathbf{0} & \mathbf{0} & \dots & 
    \mathbf{R}_m
    \end{pmatrix}^{-1}\mathbf{F}_f^{T}\mathbf{x}^{(it+1)}$
    
    $\begin{pmatrix}
    \mathbf{R}_0 & \mathbf{0} & \dots & \mathbf{0} \\
    \mathbf{0} & \mathbf{R}_1 & \cdots & \mathbf{0} \\
     \vdots & \vdots & \ddots & \vdots \\
    \mathbf{0} & \mathbf{0} & \dots & 
    \mathbf{R}_m
    \end{pmatrix}=$
    
    $=\begin{pmatrix}
    \mathbf{F}_0^{T}\mathbf{F}_0 & \mathbf{0} & \dots & \mathbf{0} \\
    \mathbf{0} & \mathbf{F}_1^{T}\mathbf{F}_1 & \cdots & \mathbf{0} \\
     \vdots & \vdots & \ddots & \vdots \\
    \mathbf{0} & \mathbf{0} & \dots & 
    \mathbf{F}_m^{T}\mathbf{F}_m
    \end{pmatrix} + $
    
    $+ \beta \begin{pmatrix}
    \mathbf{I}_d & \mathbf{0} & \dots & \mathbf{0} \\
    \mathbf{0} & \mathbf{I}_d & \cdots & \mathbf{0} \\
     \vdots & \vdots & \ddots & \vdots \\
    \mathbf{0} & \mathbf{0} & \dots & 
    \mathbf{I}_d
    \end{pmatrix}=\mathbf{F}_f^{T}\mathbf{F}_f + \beta \mathbf{I}_{dm}$
   
    \begin{equation}
        \label{equiv_step_3}
        \Rightarrow \mathbf{w}^{(it+1)}=(\mathbf{F}_f^{T}\mathbf{F}_f+\beta\mathbf{I}_{dm})^{-1}\mathbf{F}_f^{T}\mathbf{x}^{(it+1)}
    \end{equation}
    
     $\text{Step}_1^{(it+1)}$ and $\text{Step}_2^{(it+1)}$ remain unchanged if we compute $\mathbf{w}$ per frame. From Eq.~\ref{equiv_step_3} and Eq.~\ref{equiv_step_4} we conclude that computing $\mathbf{w}$ per frame results in Algorithm~\ref{alg:practical_iterations_per_frame}, which is equivalent to Algorithm~\ref{alg:practical_iterations}, where we have replaced $\mathbf{F}$ with $\mathbf{F}_f$.
     
     \begin{algorithm}
        \caption{\textcolor{blue}{GO-VOS - with $\mathbf{w}$ learned per frame}}
        \label{alg:practical_iterations_per_frame}
        \begin{algorithmic}[1]
            \Statex \textbf{Propagation:}
            \State $\mathbf{x}^{(it+1)} \gets \mathbf{M}\mathbf{x}^{(it)}$ 
            \label{practical_iterations_alg_step1}
            \State $\mathbf{x}^{(it+1)} \gets \mathbf{x}^{(it+1)} / max(\mathbf{x}^{(it+1)})$ \label{practical_iterations_alg_step2}
            \Statex \textbf{Regression:}
            \State $\mathbf{w}^{(it+1)}=(\mathbf{F}_f^{T}\mathbf{F}_f+\beta\mathbf{I}_{dm})^{-1}\mathbf{F}_f^{T}\mathbf{x}^{(it+1)}$
            \label{practical_iterations_alg_step3}
            \Statex \textbf{Projection:}
            \State $\mathbf{x}^{(it+1)} \gets \mathbf{F}_f\mathbf{w}_f^{(it+1)}$ \label{practical_iterations_alg_step4}
        \end{algorithmic}
    \end{algorithm}
    
    \hfill \quad $\square$

    Following the same procedure as in Sec.~\ref{convergence_proof}, we can prove that Algorithm~\ref{alg:practical_iterations_per_frame} is equivalent to a power iteration scheme, computing the leading eigenvector of a slightly different Feature-Motion matrix:
    
    \begin{equation}
        \label{}
        \begin{aligned}
        \mathbf{x}^{(it+1)}=\frac{\mathbf{A}_{alg}^{f}\mathbf{x}^{(it)}}{\|\mathbf{A}_{alg}^{f}\mathbf{x}^{(it)}\|_2} \\
        \text{where }A_{alg}^{f}=\mathbf{M}\mathbf{F}_f(\mathbf{F}_f^{T}\mathbf{F}_f+\beta\mathbf{I}_{dm})^{-1}\mathbf{F}_f^{T}
        \end{aligned}
    \end{equation}
    
    Also similar to the approach of Sec.~\ref{convergence_proof}, it can be proved that Algorithm~\ref{alg:practical_iterations_per_frame} solves the following problem:
    
     \begin{equation}
        \label{optimization_problem_alg_per_frame}
        \begin{aligned}
        \textbf{Problem 3:} \quad \mathbf{x}^{*} = \argmax_{\mathbf{x}} \mathbf{x}^{T}\mathbf{A}_{alg}^{f}\mathbf{x} \quad \text{s.t.} \quad \|\mathbf{x}\|_2=1 \\
        \text{where }
        \mathbf{A}_{alg}^{f}=\mathbf{M}\mathbf{F}_f(\mathbf{F}_f^{T}\mathbf{F}_f+\beta\mathbf{I}_{dm})^{-1}\mathbf{F}_f^{T}
        \end{aligned}
    \end{equation}
    
\section{Complexity analysis}\label{complexity}
   Considering our formulation, as a graph with one-to-one correspondences between nodes and video pixels, one would expect the problem to be intractable. Yet, our solution avoids actually computing the adjacency matrix $\mathbf{M}$ and the Feature-Motion matrix $\mathbf{A}_{alg}$, resulting in a fast implementation. We further perform an analysis of the complexity of each step, which is the most relevant as the actual time required for performing the computations is highly influenced by factors like: programming language and considered hardware. 
    
    \textbf{Propagation - $O(nk)$}. First step of Algorithm 1, $\mathbf{x}^{(it+1)}=\mathbf{M}\mathbf{x}^{(it)}$, can be written for a node $i$ as $x_i \gets \sum_j \mathbf{M}_{ij}x_j$. We implement this efficiently propagating the soft labels $x_i$, weighted by the pairwise terms $\mathbf{M}_{ij}$, to all the nodes from the other frames to which $i$ is connected in the graph (Sec. 3 in our paper). We consider only a diameter of $k$ frames around the frame associated to node $i$, as $\mathbf{M}_{ij}=k(i,j)$ decreases rapidly towards zero with the temporal distance between $i$ and $j$. In consequence, the complexity of this step is $O(nk)$, where $k=11$ in our experiments. 
    
    \textbf{Regression - $O(d^2n)$}. Step 3 of Algorithm 1 will compute $\mathbf{w}^{(it+1)}=(\mathbf{F}^{T}\mathbf{F}+\beta\mathbf{I}_d)^{-1}\mathbf{F}^{T}\mathbf{x}^{(it+1)}$. The complexity of $\mathbf{F}^{T}\mathbf{F}$ is $O(d^2n)$, the complexity of $\mathbf{F}^{T}\mathbf{x}^{(it+1)}$ is $O(dn)$ and the complexity of computing the inverse $(\mathbf{F}^{T}\mathbf{F}+\beta\mathbf{I}_d)^{-1}$ is $O(d^3)$. As $n$ is the number of video pixels, $n\gg d$ and $O(d^2n)$ asymptotically dominates $O(dn)$ and $O(d^3)$. In consequence, the complexity of Step 2 is $O(d^2n)$. We highlight that in all our experiments $d\leq56$ and it takes only $0.009$ sec/frame to compute the inverse $(\mathbf{F}^{T}\mathbf{F}+\beta\mathbf{I}_d)^{-1}$ and it would require only $0.1$ sec/frame in case we raise $d$ to consider thousands of features (e.g.  $2048$). Another important aspect is that $\mathbf{P}=(\mathbf{F}^{T}\mathbf{F}+\beta\mathbf{I}_d)^{-1}\mathbf{F}^{T}$ should be computed only once, and further used in each iteration of GO-VOS.
    
    \textbf{Projection - $O(dn)$}. The last step of Algorithm 1, $\mathbf{x}^{(it+1)} \gets \mathbf{F}\mathbf{w}^{(it+1)}$ has a complexity $O(dn)$.
    
    The complexity of the regression step asymptotically dominates all other complexities, resulting in a complexity $O(d^2n)$ for GO-VOS.
    
    As mentioned, various information can be computed only once, and reused in each iteration of GO-VOS. Finally, our implementation requires $0.04$ sec/frame for computing optical flow using FlowNet2.0 \cite{ilg2017flownet}, $0.17$ sec/frame in order to precompute information for propagation and regression steps and $0.4$ sec/frame for the optimization process. We report the runtime for our PyTorch implementation, on a computer with the following specifications: Intel(R) Xeon(R) CPU E5-2697A v4 @ 2.60GHz, GPU GeForce GTX 1080.

\section{Quantitative results - YouTube-Objects v1.0}\label{quantitative_res}
    In Table~\ref{tab:experiments_yto_results} we introduce quantitative results of GO-VOS on v1.0 of YouTube-Objects dataset, following the same experimental setup as in Sec 4.3 of our original paper. GO-VOS is the best for 7 out of 10 classes and in average.
    \begin{table}[h]
        \begin{center}
        \resizebox{\columnwidth}{!}{%
        \begin{tabular}{|c||c|c|c|c|c|c|c|c|c|c||c||c|}
            \hline
            Method & aero & bird & boat & car & cat & cow & dog & horse & moto & train & avg & sec/frame\\
            \hline\hline
            \cite{prest2012learning} & 51.7 & 17.5 & 34.4 & 34.7 & 22.3 & 17.9 & 13.5 & 26.7 & 41.2 & 25.0 & 28.5 & N/A \\ 
            \hline
            \cite{papazoglou2013fast} & 65.4 & 67.3 & 38.9 & 65.2 & 46.3 & 40.2 & 65.3 & 48.4 & 39.0 & 25.0 & 50.1 & 4 \\ 
            \hline
            \cite{zhang2015semantic} & 75.8 & 60.8 & 43.7 & 71.1 & 46.5 & 54.6 & 55.5 & 54.9 & 42.4 & 35.8 & 54.1 & N/A \\ 
            \hline
            \cite{jun2016pod} & 64.3 & 63.2 & 73.3 & 68.9 & 44.4 & 62.5 & 71.4 & 52.3 & 78.6 & 23.1 & 60.2 & N/A \\ 
            \hline
            HPP\cite{haller2017unsupervised} & 76.3 & 71.4 & 65.0 & 58.9 & 68.0 & 55.9 & 70.6 & 33.3 & 69.7 & 42.4 & 61.1 & 0.35 \\
            \hline
            \cite{croitoru2017unsupervised} & 77.0 & 67.5 & \textbf{\textcolor{blue}{77.2}} & 68.4 & 54.5 & \textbf{\textcolor{blue}{68.3}} & 72.0 & \textbf{\textcolor{blue}{56.7}} & 44.1 & 34.9 & 62.1 & 0.04 \\
            \hline\hline
            \shortstack[]{\textcolor{blue}{\textbf{GO-VOS}} \\ \textcolor{blue}{\textbf{unsupervised}}} & \textbf{\textcolor{blue}{88.2}} & \textbf{\textcolor{blue}{82.5}} & 62.7 & \textbf{\textcolor{blue}{76.7}} & \textbf{\textcolor{blue}{70.9}} & 50.0 & \textbf{\textcolor{blue}{81.9}} & 51.8 & \textbf{\textcolor{blue}{86.2}} & \textbf{\textcolor{blue}{55.8}} & \textbf{\textcolor{blue}{70.7}} & 0.91 \\
            \hline
        \end{tabular}
        }
        \end{center}
        \caption{Quantitative results of our method (CorLoc metric), compared with state of the art solutions on YouTube-Objects v1.0 dataset. We have state of the art results for 7 out of 10 classes, and in average. \cite{croitoru2017unsupervised} is much faster as it is based on a trained network and requires only a forward pass at test time. (blue bold - best solution)}
        \label{tab:experiments_yto_results}
        \vspace{-6mm}
    \end{table}

\section{Qualitative results}\label{qualitative_res}
\begin{figure*}[!h]    
        \includegraphics[width=\textwidth]{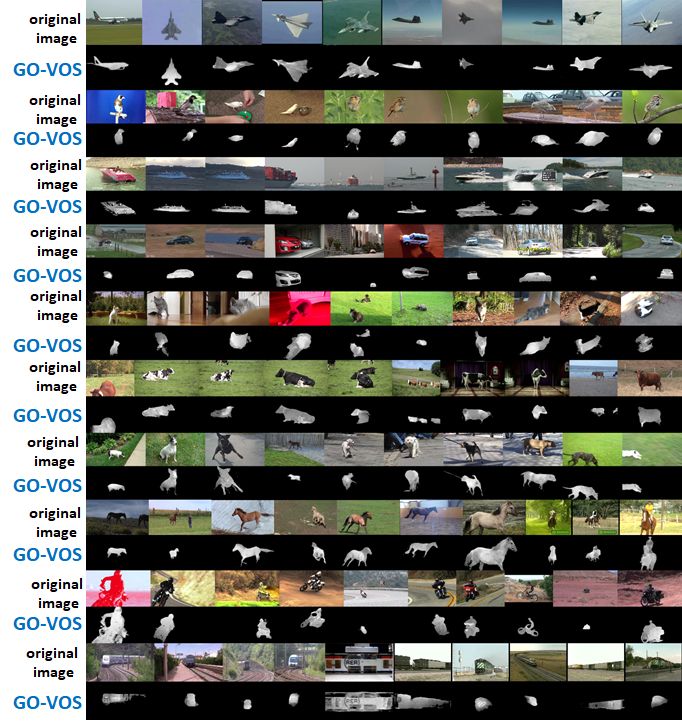}
        \centering
        \caption{Qualitative results on YouTube-Objects dataset.}
        \label{fig:yto_results}
     
\end{figure*}

\begin{figure*}[!h]    
        \includegraphics[width=\textwidth]{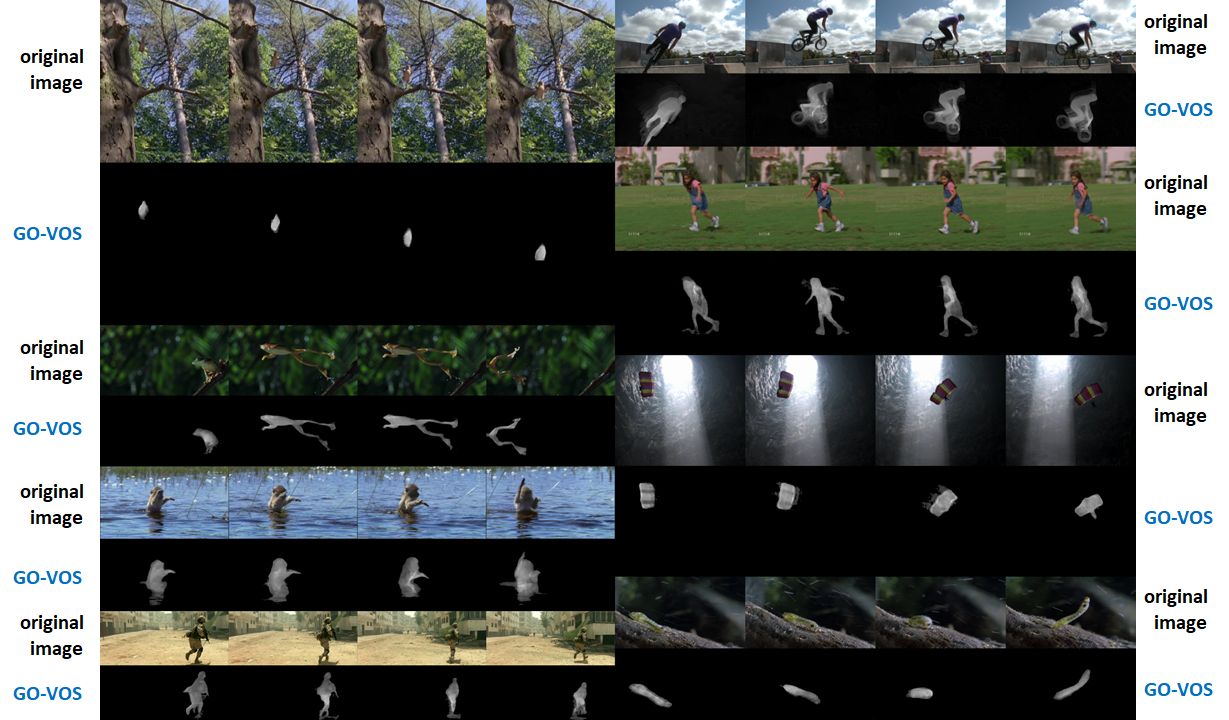}
        \centering
        \caption{Qualitative results on SegTrack dataset.}
        \label{fig:segtrack_results}
\end{figure*}

    In Figure~\ref{fig:yto_results} and Figure~\ref{fig:segtrack_results} we introduce qualitative results of our method on YouTube-Objects dataset and SegTrack dataset. For SegTrack we present results for 8 videos, while for YouTube-Objects we have selected 10 samples for each class. In video \textbf{288\textunderscore qualitative\textunderscore results.avi} we present qualitative results of our method on DAVIS dataset.

    We also introduce three video sequences that will help us highlight key aspects of our solution and to compare it with other state of the art methods.
    
    \textbf{Qualitative evolution of soft-segmentation masks:} In the first video (\textbf{288\textunderscore qualitative\textunderscore evolution.avi}) we present the evolution of the soft-segmentation masks over several iterations of our algorithm. For this examples, GO-VOS is initialized with non-informative soft-segmentation masks (random masks) and we consider only unsupervised features (motion cues). We present few videos and for each we choose 6 representative frames that are observed over several iterations. This examples highlight how the object of interest emerges from an initially random mask. With every iteration, the soft-segmentation masks become clearer and the foreground region is accentuated.  
    
    \textbf{The role of features:} In the second video (\textbf{288\textunderscore the\textunderscore role\textunderscore of\textunderscore features.avi}), we visually prove that the convergence point of GO-VOS is strictly determined by the Feature-Motion matrix $\mathbf{A}_{alg}$, dependent of the spacetime graph structure (matrix $\mathbf{M}$) and the set of chosen features (matrix $\mathbf{F}$). For this purpose, we define four configurations for GO-VOS, varying the initialization masks and the set of features. We either use informative initialization: the soft-segmentation masks provided by \cite{song2018pyramid}; or non-informative initialization: a central Gaussian. Regarding the features, we use unsupervised features: motion cues along the optical flow chains; or supervised features: in addition to motion cues we also use the predictions of \cite{song2018pyramid} as features. We want to prove that solutions with the same set of features will converge to the same solution irrespective of the initialization. In the provided video it can be observed that with each iteration, the soft-segmentation masks provided by unsupervised solutions are becoming more and more similar, and the same behaviour is observed for the ones generated by supervised solutions, although they start from different initializations. 
    
    \textbf{Qualitative comparison:} In the last video (\textbf{288\textunderscore qualitative\textunderscore comparison.avi}) we compare the results of GO-VOS with other state of the art solutions on several videos. For comparison purpose we have chosen two supervised solutions (supervised training for the object segmentation task): PDB \cite{song2018pyramid} and ARP \cite{koh2017primary}; and two unsupervised solutions (no supervised training for the  object segmentation task): ELM \cite{lao2018extending} and FST \cite{papazoglou2013fast}. For this test we have considered the unsupervised formulation of GO-VOS as we only use unsupervised features extracted along the optical flow chains: motion and color cues. Regarding the initialization, for this test, GO-VOS is initialized with a non-informative central Gaussian. We emphasize that in multiple sequences GO-VOS overcomes both supervised and unsupervised solutions.
    
\section{Conclusions}\label{conclusions}
    We have provided formal proofs that the proposed algorithm is guaranteed to converge to the leading eigenvector of the Feature-Motion matrix $\mathbf{A}_{alg}$ and that this convergence point is also the global optimum of our objective: Problem 2. Additionally, we prove that under the modified setup, where $\mathbf{w}$ is computed per frame, we arrive at an algorithm with the same properties as the initial algorithm, guaranteed to converge and to optimally solve a slightly different problem: Problem 3, that differs from Problem 2 in the structure of features matrix $\mathbf{F}$.
    We have strengthened the position of our algorithm, presenting state of the art results for v1.0 of YouTube-Objects dataset, qualitative results on both YouTube-Objects and SegTrack datasets, along with a detailed complexity analysis of GO-VOS. 

{\small
\bibliographystyle{ieee}
\bibliography{bib}
}   

\clearpage
\clearpage
